\documentclass{article}
\usepackage{microtype}
\usepackage{graphicx}
\usepackage{wrapfig}
\usepackage{subcaption}
\usepackage{booktabs} 
\usepackage{multirow}
\usepackage{hyperref}
\usepackage{tabularx}
\usepackage{float}      
\usepackage{placeins}   

\usepackage{xcolor}
\usepackage{amssymb}

\PassOptionsToPackage{numbers, compress}{natbib}

\usepackage[final,main]{neurips_2026}

\usepackage{amsmath}
\usepackage{amssymb}
\usepackage{mathtools}
\usepackage{amsthm}
\usepackage{bbm}
\newcolumntype{Y}{>{\centering\arraybackslash}X}

\usepackage{listings}

\usepackage[capitalize,noabbrev]{cleveref}
\theoremstyle{plain}

\theoremstyle{definition}

\theoremstyle{remark}

\usepackage{xspace}

\newcommand{\methodname}{Monroe\xspace}

\newcommand{\minimolpp}{\ensuremath{\text{MiniMol}_{\mathsf{PFN}}}\xspace}
\newcommand{\chemeleonpp}{\ensuremath{\text{CheMeleon}_{\mathsf{PFN}}}\xspace}
\newcommand{\methodnameens}[1]{\ensuremath{\text{Monroe}_{\mathsf{ENS}}(#1)}}
\DeclareMathOperator*{\argmin}{argmin}

\newif\ifsd
\sdtrue

\begin{document}

\title{\methodname: A Molecular Foundation Model\\for In-Context Probabilistic Inference}

\author{%
  B{\l}a\.zej Banaszewski\\
  \texttt{blazej@banaszewski.pl}\\
  Graphcore\\
  Max Planck Institute of Biochemistry
  \And
  Andrew W. Fitzgibbon\\
  \texttt{awf@fitzgibbon.ie}\\
  Graphcore
}

\maketitle

\begin{abstract}
Bioassay activity prediction is often data-limited because drug-discovery datasets rely on time-consuming and expensive wet-lab experiments for data generation and evaluation.
This challenge has inspired recent research into molecular foundation models (MFMs), which aim to encode general-purpose chemical knowledge into molecular representations that generalize well in data-constrained scenarios. This paper presents \methodname, a new MFM with several innovations over the existing state of the art: increased scale allowing pre-training on over 81 million molecules from the PM6 quantum chemistry dataset; improved graph representation of stereochemistry; improved training losses including conformer denoising and embedding decorrelation; improved multi-task learning; and the use of a prior-data-fitted model (TabPFNv3) for downstream in-context prediction. Our evaluations use a principled pairwise comparison framework that measures statistically significant performance differences. Across established Polaris benchmarks, \methodname matches or exceeds existing MFMs, while on activity cliff benchmarks, designed to assess utility for molecular discovery, it achieves significant improvements over prior methods. Finally, ablation and transfer experiments show that PFN-based downstream predictors also substantially improve two leading existing models, MiniMol and CheMeleon, yielding new state-of-the-art variants we call \minimolpp and \chemeleonpp, suggesting that our downstream adaptation strategy generalizes beyond \methodname.
Source code is at \href{https://github.com/blazejba/monroe}{github.com/blazejba/monroe}.

\end{abstract}

\section{Introduction}

\begin{figure}[h]
    \centering
    \makebox[\textwidth][c]{%
        \includegraphics[width=1.00\textwidth]{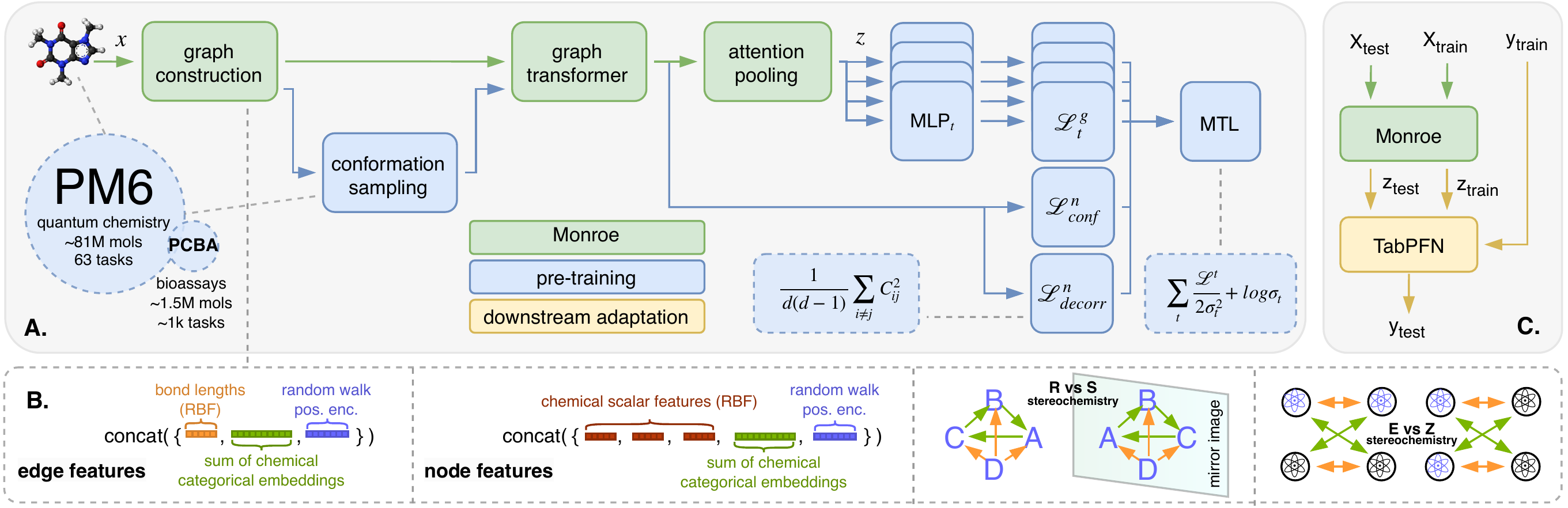}%
    }
    \caption{A. Monroe multi-task pretraining with structural and decorrelation losses under uncertainty weighting. B. Graph construction including featurization and stereochemistry-aware rewiring. C. In-context probabilistic inference: embeddings generated by attention pooling of the pretrained model together with training labels are passed to TabPFN, which outputs an approximate posterior-predictive distribution for the test molecule in a single forward pass.}
    \label{fig:architecture}
\end{figure}

\textit{In silico} biological property prediction for small molecules is typically performed using wet-lab bioassay datasets whose scales are inherently limited, as biological experiments are expensive, time-consuming, and can only be conducted at scale in a small number of specialized facilities.
Traditional machine learning approaches fit a task-specific, low-capacity model directly to small datasets, while more recent approaches seek to learn molecular representations that transfer across tasks, enabling accurate prediction in data-scarce settings.
This paradigm, often referred to as \emph{molecular foundation models} (MFMs), aims to extract generalizable chemical and structural knowledge from large, readily available data sources, which can then be adapted to downstream biological tasks.
In drug discovery, this is particularly appealing for applications such as virtual screening, where models are repeatedly used to prioritize compounds before committing to further rounds of costly experimental evaluation.

In this paper, we introduce a new MFM, called \methodname,\!\NoHyper\footnote
{Named after the computational quantum chemistry pioneer Elizabeth Monroe Boggs.}
based on a number of innovations over the current state of the art:
1. improved multitask learning using uncertainty-weighted (UW) loss balancing;
2. improved training losses including conformer denoising and embedding decorrelation;
3. improved graph representation of stereochemistry;
4. replacing standard fine-tuning with downstream in-context prediction using a prior-data-fitted model;
and 5. scaling training to allow pretraining on the PM6 dataset, comprising semi-empirical quantum properties and quantum chemistry (QC) conformers, and the PCBA dataset, consisting of high-throughput bioassay activity readouts.

We follow recent best practices in model evaluation~\cite{walters2024polaris} to make statistically sound comparisons between models, and show that this combination of innovations produces a model which exceeds state-of-the-art performance on standard Polaris benchmarks, and on newer activity-cliff benchmarks
~\cite{vanTilborg2022moleculeace}. In addition to standard ablations over the design choices, we also augment existing methods with TabPFN-based downstream prediction and obtain two new state-of-the-art variants: \minimolpp (augmenting MiniMol~\cite{klaser2024minimol}) and \chemeleonpp (augmenting CheMeleon~\cite{burns2025chemeleon}). \minimolpp is the closest competitor to \methodname on Polaris but significantly weaker on MoleculeACE, and the opposite is true for \chemeleonpp, which is significantly behind on Polaris and closer on MoleculeACE.

\section{Related Work}

\paragraph{Molecular property prediction.}
Training objectives for pretraining molecular foundation models span a wide range of approaches:
unsupervised \citep{mendezlucio2024mole,ross2022molformer}, multi-task \citep{sypetkowski2024molgps,klaser2024minimol,burns2025chemeleon}, and hybrids that combine molecular property prediction with denoising \citep{lu2023unimolplus}. Relevant training data sources used by these methods cover a wide range of modalities: bioassays \citep{trannguyen2020litpcba}, quantum chemical properties \citep{Nakata2020PubChemQC}, geometrical and molecular descriptors \citep{moriwaki2018mordred} and proprietary data like phenomics \citep{sypetkowski2024molgps}. Architectures vary as well, including SMILES language models \citep{ross2022molformer}, graph transformers \citep{ying2021do},
and message-passing networks \citep{yang2019chemprop,klaser2024minimol,sypetkowski2024molgps}.
Our approach derives from GRIT \citep{pmlr-v202-ma23c}, which augments a transformer with a range of inductive biases developed in graph learning, including random-walk positional encodings \citep{dwivedi2022graph},
graph attention mechanisms \citep{velivckovic2018gat},
attention-based global pooling \citep{li2016gated,ying2018hierarchical},
and the use of a virtual node for global information propagation \citep{gilmer2017neural}.
We rely on TabPFN \citep{hollmann2025tabpfn}, a prior-data-fitted model that performs amortized in-context prediction and can be interpreted as approximating Bayesian posterior-predictive inference under its training prior.

\paragraph{Multi-task learning.}
Previous MFM training has cast the problem as multi-task learning (MTL), but has typically used only the simplest MTL methods.  The wide variety of task types and losses used in this work requires deeper consideration of MTL techniques, beyond Equal Weighting (EW), namely
Uncertainty Weighting (UW) \citep{kendall2018uncertainty},
Dynamic Weight Average (DWA) \citep{liu2019endtoend},
Random Loss Weighting (RLW) \citep{lin2022rlw},
and Smooth Tchebycheff (STCH) \citep{lin2022smooth}.
Beyond these {\em scalarization} methods are gradient-based methods that manipulate per-task gradients, but which require $T$ backward passes for $T$ tasks, which becomes prohibitive for $T$ of the order of a thousand as in this work.

\paragraph{Conformational prediction.} Uni-Mol+ \citep{lu2023unimolplus} couples conformation
refinement with QC property prediction: it starts from a low-cost RDKit conformer,
iteratively updates coordinates toward a DFT equilibrium structure, and predicts
properties from the refined geometry. To reduce the train--test gap, it samples
intermediate conformers along a pseudo-linear trajectory between the raw and
equilibrium conformations during training. LoQI \citep{nikitin2025loqi} targets
stereochemistry explicitly by augmenting molecular graphs with auxiliary edges for
E/Z double bonds and R/S chiral centers, enabling stereochemistry-aware generation.

In our work, we leverage the strengths of both of these approaches, namely:
conformation refinement and stereochemistry-awareness, and we explain why
the latter is important despite the apparently present stereochemically-disambiguating
structural information. We do not adopt an SE(3)-equivariant backbone, reasoning that benchmarks such as TDC \citep{Huang2021tdc} are dominated by non-equivariant models, and sequence-only methods like MolFormer \citep{ross2022molformer} remain competitive on the same Polaris tasks we evaluate on, indicating that coordinate-level equivariance is not a prerequisite for strong downstream performance.
Stereochemistry, however, clearly matters: \citet{nikitin2025loqi} report that an SE(3)-equivariant baseline distinguishes R/S configurations only 38\% of the time without auxiliary stereo edges, rising to 96\% once they are added---coordinates alone are not sufficient despite the architecture in principle handling handedness. Coordinate-level equivariance is therefore neither necessary for our downstream targets nor sufficient for stereochemical disambiguation, motivating our choice of an invariant encoder augmented with explicit stereo edges.

\section{Methods}
\def\vv#1{\mathbf{#1}}
\def\vxi{\vv x_i}
\def\vxj{\vv x_j}
\def\eij{\vv e_{ij}}
\def\sijh{\vv s_{ij}^h}
\def\aijh{{\boldsymbol \alpha}_{ij}^h}
\def\mih{\vv m_i^h}
\def\fpsi{f_\psi}

\methodname follows established multi-task pretraining practice~\cite{Beaini2024Towards,sypetkowski2024molgps,klaser2024minimol,burns2025chemeleon}.
As illustrated in Figure~\ref{fig:architecture}, the architecture comprises a trunk $\fpsi$, which is applied to input graph $G$ to produce a ``fingerprint'' vector $\vv z = \fpsi(G)$.
Each pre-training task $t$ has a dedicated lightweight MLP head on top of the shared encoder. These heads are used only for the pre-training objectives and are discarded thereafter, retaining only the encoder $f_\psi$ for downstream tasks. Pretraining also includes a node-level task of structure denoising, explained later.

\subsection{Architecture}
We build on GRIT \citep{pmlr-v202-ma23c}, a graph transformer that augments self-attention with random-walk positional encodings and per-edge keys and values; the full layer update is given in \cref{sec:architectural-details}. Our variant differs from the original in three ways. First, attention is evaluated only on observed edges, rather than on the full $|V|\times|V|$ pair set, and the base graph is symmetrized so message passing is undirected. This yields sparse, $O(|E|)$ attention that scales to molecular graphs. Second, for graph-level representations we use multi-head attention pooling \citep{lee2019set}, where each head learns to weight nodes differently before aggregation. Third, a virtual node connected to all atoms enables global all-to-all information exchange at each layer, which has been shown to be valuable for accurate structure prediction~\citep{jumper2021alphafold,lu2023unimolplus}.

\paragraph{Featurization.}
\def\vfi{\vv f_i}
\def\vp{\vv p}
\def\vpi{\vp_i}
\def\vpj{\vp_j}
Node and edge features combine categorical and scalar chemical features from RDKit,
input conformers from RDKit,
and random-walk positional encodings (see Figure \ref{fig:architecture}B).

\def\knode{K_{\mathrm{node}}}
\def\fnode{F_{\mathrm{node}}}
\def\embnode{\mathrm{Emb}^\mathrm{node}}
Let a molecule be a graph $G=(V,E)$.
For each node $i\in V$,
let $\{c_{i,k}\}_{k=1}^{\knode}$ denote $\knode$ categorical codes,
$\vfi\in\mathbb{R}^{\fnode}$ continuous RDKit features,
and $\vpi\in\mathbb{R}^3$ atom coordinates.
Let $P\in\mathbb{R}^{|V|\times|V|}$ be the random-walk transition matrix on the undirected graph with self-loops, and let $R$ be the maximum walk length.
Using embedding tables $\embnode_k:\mathcal{C}_k\to\mathbb{R}^{d_{\mathrm{emb}}}$, we form
\begin{equation}
\begin{aligned}
\vv u_i = \sum_{k=1}^{\knode} \embnode_k(c_{i,k}) 
, \quad
\vv r_i = \big[(P^1)_{ii},\ldots,(P^R)_{ii}\big] \in \mathbb{R}^{R}, \quad
\vxi^{(0)} = W_I\,[\vv u_i \,\|\, \vfi \,\|\, \vv r_i] + W_{Ib},
\end{aligned}
\end{equation}
where $W_I\in\mathbb{R}^{d_h\times(d_{\mathrm{emb}}+\fnode+R)}$ and $W_{Ib}\in\mathbb{R}^{d_h}$ project the concatenated input to the encoder hidden dimension $d_h$.

\def\kedge{K_{\mathrm{edge}}}
\def\vphiij{\boldsymbol{\phi}_{ij}}
Edges use categorical bond features together with radial basis encodings of bond
lengths $d_{ij}=\|\vpi-\vpj\|_2$ from the input conformer:
\begin{equation}
\phi_m(d_{ij}) = \exp\!\big(-\gamma(d_{ij}-\mu_m)^2\big), \qquad
\vphiij = \big[\phi_1(d_{ij}),\ldots,\phi_M(d_{ij})\big] \in \mathbb{R}^M.
\end{equation}
Here $M$ is the number of radial basis functions, $\{\mu_m\}_{m=1}^{M}$ are fixed RBF centers (uniformly spaced over the bond-length range),
and $\gamma$ controls the bandwidth; we set $\gamma$ based on the spacing between
centers so each basis function covers a comparable length scale. Combining with random walk and edge-categoricals:
\begin{equation}
\begin{aligned}
r_{ij} = \tfrac12\Big[
P^1_{ij}+P^1_{ji}, \ldots, P^R_{ij}+P^R_{ji}
\Big], \quad
u_{ij} =\sum_{k=1}^{\kedge} \mathrm{Emb}^e_k(c_{ij,k}), \quad
\eij^{(0)} = W_e\,[u_{ij} \,\|\, \vphiij \,\|\, r_{ij}] + b_e ,
\end{aligned}
\end{equation}
with $\kedge$ edge categorical codes ($\kedge=4$), $W_e\in\mathbb{R}^{d_e\times(d_{\mathrm{emb}}+M+R)}$ and $b_e\in\mathbb{R}^{d_e}$ projecting to the edge hidden dimension $d_e$. For more details on featurization see Appendix \ref{sec:featurization-details}.

\paragraph{Stereochemistry-aware representation.}
\label{sec:stereo-aware}
Our encoder consumes only rotation- and reflection-invariant features (bond lengths
and graph attributes). This design choice, while ensuring geometric consistency,
creates a fundamental problem: \emph{enantiomers} (mirror-image molecules) and
\emph{E/Z isomers} (geometric isomers around double bonds) can collapse to identical
inputs, despite having drastically different biological behaviors. Following
LoQI \citep{nikitin2025loqi}, we augment the molecular graph with auxiliary
edges that encode E/Z and R/S configurations (see \cref{sec:stereo-details} for background).

\begin{figure}
\centering
  \includegraphics[width=1.0\linewidth]{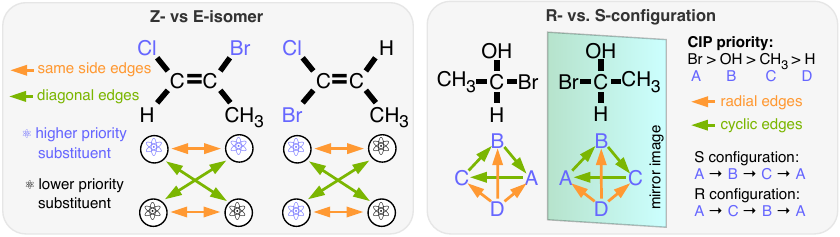}
\caption{Stereochemical E/Z isomers and R/S configurations.}
\label{fig:stereo_configs}
\vspace{0em}
\end{figure}

For each double bond with substituents prioritized by CIP rules, let
$(s_1,s_2)$ denote the higher-priority substituent on each carbon and
$(t_1,t_2)$ the lower-priority ones. We add ``parallel'' edges connecting
same-side substituents and ``diagonal'' edges connecting opposite-side substituents:

\begin{equation}
\mathcal{E}_{\mathrm{EZ}}=\{(s_1,s_2),(t_1,t_2)\}, \quad
\mathcal{E}_{\mathrm{EZ}}^{\perp}=\{(s_1,t_2),(t_1,s_2)\}.
\end{equation}

The observed E/Z label is attached to $\mathcal{E}_{\mathrm{EZ}}$ and
its opposite to $\mathcal{E}_{\mathrm{EZ}}^{\perp}$, ensuring the message-passing
network can distinguish both isomers regardless of input ordering (see \Cref{fig:stereo_configs}A).

For a tetrahedral stereocenter with CIP-ordered neighbors $(a,b,c,d)$
(highest to lowest priority), we add edges that encode the chiral arrangement:
\begin{equation}
\mathcal{E}_{\mathrm{RS}}=\underbrace{\{(a,d),(b,d),(c,d)\}}_{\text{radial}}
\cup \underbrace{\{(a,b),(b,c),(c,a)\}}_{\text{cyclic (S; reversed for R)}}.
\end{equation}
The radial edges connect the lowest-priority substituent $d$ to all others.
The cyclic edges follow the $a{\to}b{\to}c{\to}a$ direction, which is counterclockwise when viewed from $d$ and therefore encodes the S configuration; the reversed cycle $a{\to}c{\to}b{\to}a$ encodes R. Flipping the cycle direction in the graph encoding distinguishes enantiomers (see \Cref{fig:stereo_configs}B).

\subsection{Training}
\label{sec:methods-training}

\paragraph{Task weighting.}
We balance task losses $\{\mathcal{L}_t\}_{t=1}^{T}$ via Uncertainty Weighting \citep{kendall2018uncertainty}, which assigns each task a learnable standard deviation $\sigma_t > 0$ and minimizes
\begin{equation}
\mathcal{L}_{\mathrm{UW}} = \sum_{t=1}^{T}
\left[
\frac{\mathcal{L}_t}{2\sigma_t^2} + \log \sigma_t
\right].
\end{equation}
Tasks with high observation noise (large $\sigma_t$) are downweighted automatically via the precision term, while the $\log \sigma_t$ regularizer prevents the trivial solution $\sigma_t \!\to\! \infty$.
For numerical stability, we parameterize $\sigma_t = e^{s_t/2}$,
giving the equivalent objective
 $\tfrac12 \sum_t [e^{-s_t} \mathcal{L}_t + s_t]$
and we learn $\{s_t\}$ jointly with the encoder.

\paragraph{Conformer denoising and augmentation.}
\label{sec:sd}
Inspired by Uni-Mol+ \citep{lu2023unimolplus}, we train the model to recover high-quality equilibrium structures from noisy estimates (see Figure \ref{fig:architecture}A).
Input positions $\vp^{\mathrm{in}}$ are sampled by interpolating between a reference $\vp^{\mathrm{ref}}$ (PM6-optimized) and a low-cost approximation $\vp^{\mathrm{rdk}}$, generated using RDKit's ETKDGv3 \citep{landrum2016rdkit} followed by MMFF94s force field minimization:
$\vpi^{\mathrm{in}} = \eta\,\vpi^{\mathrm{ref}} + (1-\eta)\,\vpi^{\mathrm{rdk}}$.
We use $\eta=1$ with probability $0.1$, $\eta=0$ with probability $0.8$, and $\eta\sim\mathcal{U}(0.4,0.6)$ with probability $0.1$.

\def\predpi{\hat{\vp}_i}
The model predicts atomic coordinates $\predpi \in \mathbb{R}^3$ from last-layer node representations $\vxi^{(L)}$.
To handle the rotational and translational invariance of molecular systems, we calculate the loss based on the ``best fit'' alignment.
\def\Rstar{R^{*}}
\def\tstar{t^{*}}
We analytically compute the optimal rotation matrix $\Rstar$ and translation vector $\tstar$ that align the predicted coordinates to the reference structure $\vp^{\mathrm{ref}}$ by minimizing the $L_2$ error~\citep{kabsch1976solution}.
The loss is then computed on the aligned coordinates:
\begin{equation}
\mathcal{L}_{\mathrm{conf}} = \frac{1}{|V|} \sum_{i=1}^{|V|} \sum_{c=1}^{3} \ell_{\delta}\!\left( \big[(\Rstar \predpi + \tstar) - \vp^{\mathrm{ref}}_i\big]_c \right),
\quad
\Rstar, \tstar = \argmin_{R \in \mathrm{SO}(3),\, t \in \mathbb{R}^3} \sum_{i=1}^{|V|} \left\| (R \predpi + t) - \vp^{\mathrm{ref}}_i \right\|_2^2,
\end{equation}
where $\ell_{\delta}$ is the (scalar) Huber loss applied per coordinate component $c\in\{x,y,z\}$. This approach provides a more direct supervision signal than internal distance matrices while maintaining invariance to the prediction reference frame.

\paragraph{Embedding decorrelation.}
\label{sec:decorr}
The pooled graph fingerprint $\vv z \in \mathbb{R}^d$ feeds all $T$ task heads as well as the downstream TabPFN predictor. In order to make the embeddings more
tabular-like, we add a covariance penalty that drives the off-diagonal entries of the batch fingerprint covariance to zero, in the spirit of VICReg \citep{bardes2022vicreg}. For a batch of fingerprints $Z \in \mathbb{R}^{B \times d}$ centered to zero mean, with sample covariance $C = (B-1)^{-1} Z^\top Z$, the loss is
\begin{equation}
\mathcal{L}_{\mathrm{decorr}} = \frac{1}{d(d-1)} \sum_{i \neq j} C_{ij}^2,
\end{equation}
i.e., the mean squared off-diagonal covariance. This differs from the original VICReg normalization of $1/d$ by an extra factor of $1/(d-1)$, which only rescales the effective weight $w$. We add $w\,\mathcal{L}_{\mathrm{decorr}}$ to $\mathcal{L}_{\mathrm{UW}}$ outside the UW scalarization, since it acts as a regularizer.

\paragraph{Downstream adaptation.}
We use the pre-trained encoder $f_{\psi}$ as a fixed feature extractor (see Figure \ref{fig:architecture}C). For a downstream task with training data $\mathcal{D}_{\mathrm{train}}^{\mathrm{raw}} = \{(G_i, y_i)\}$,
we compute embeddings $z_i = f_{\psi}(G_i) \in \mathbb{R}^d$ to form the feature support set $\mathcal{D}_{\mathrm{train}} = \{(z_i, y_i)\}$.
These embeddings are passed to TabPFN, a prior-data-fitted model trained to approximate posterior-predictive inference on tabular datasets.
Given a test molecule $G_{\mathrm{test}}$ with embedding $z_{\mathrm{test}} = f_{\psi}(G_{\mathrm{test}})$, TabPFN outputs an approximate posterior-predictive distribution $\tilde{q}(y \mid z_{\mathrm{test}}, \mathcal{D}_{\mathrm{train}})$ in a single forward pass.
Since TabPFN is pretrained, this is amortized in-context prediction rather than explicit Bayesian inference performed at test time.
This allows for efficient in-context learning: the model adapts to the task solely by conditioning on $\mathcal{D}_{\mathrm{train}}$, eliminating the need for gradient-based fine-tuning or hyperparameter optimization (see Figure \ref{fig:architecture}C).

\paragraph{Datasets.}
We train on two complementary data sources. The first is the PM6 dataset \citep{Nakata2020PubChemQC} with labels from \citet{Beaini2024Towards}, which spans roughly 81 million molecules (see \cref{sec:pm6-details}); each molecule supplies 62 graph-level semi-empirical quantum-chemical properties, plus the conformer denoising objective described in~\S\ref{sec:sd}. The second is PCBA \citep{Beaini2024Towards}, a panel of binary PubChem bioassays covering 1.56\,M molecules: while PM6 supplies dense, large-scale physical signal, PCBA injects sparse but biologically aligned supervision closer in distribution to the downstream Polaris and MoleculeACE tasks, which the ablation in Section~\ref{sec:ablations} shows is essential for the model's downstream performance. We thus train on $T=1152$ tasks: 62 PM6 graph-level targets, 1089 PCBA per-assay binary classifications (see PCBA head below), and 1 conformer denoising objective.
Each task loss $\mathcal{L}_t$ is chosen to match the target's domain and natural error structure: we employ Huber loss for signed real-valued quantities, log-Huber for strictly positive scale-like descriptors, Beta NLL for bounded ratios, ordinal cross-entropy for integer counts (see \cref{subsec:pm6-losses} for full details), and focal binary cross-entropy for PCBA bioassays.

We filter PCBA assays to those with at least 100 positive, 100 negative, and 1000 total non-NaN training labels, retaining 1089 of the original 1328 assays (per-assay statistics and the cutoffs are shown in \cref{fig:pcba_assay_distributions}). The PCBA head is a shared linear probe (a single $\mathrm{Linear}(d, K)$ layer with $K=1089$ logits, one per surviving assay), but each assay is registered as its own task with its own learnable noise scale $\sigma_t$, so PCBA contributes $K$ independent terms to the UW sum rather than one. Per-assay class imbalance is handled by focal loss \citep{lin2017focal} with $\gamma=2$, $\alpha=0.25$.

We remove, by InChIKey, all molecules from the pretraining set that overlap with the test set of \emph{any} dataset used for downstream evaluation.
Although there is no chance of label leakage, as the tasks differ, we want to avoid a situation where any of the pretraining labels happens to be closely correlated with a downstream one.

Implementation details are given in \cref{app:code}.
Pretraining took 26 hours on four H100 GPUs.

\begin{figure*}[t]
\centering
\includegraphics[width=\textwidth]{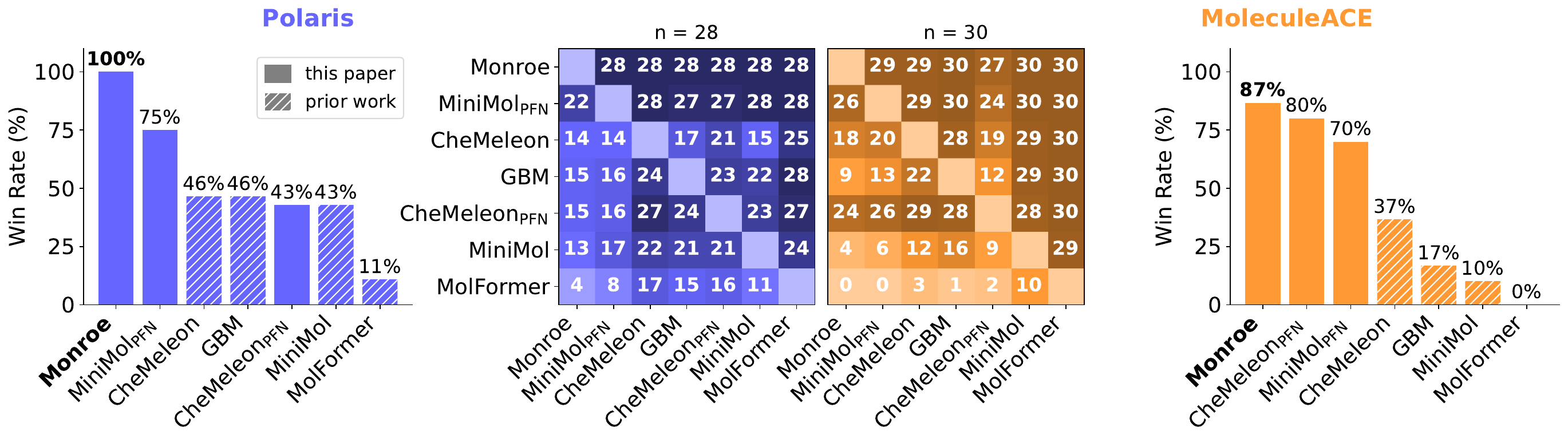}
\caption{Comparison between \{\methodname, \minimolpp, \chemeleonpp\} (methods new to this paper) and \{MiniMol, CheMeleon, MolFormer\}, the leading previous models, plus a classical QSAR, on 28 tasks from Polaris (left) and 30 tasks from MoleculeACE (right). The pairwise matrices in the middle show the number of tasks that the model on the y-axis won against the model on the x-axis. The barplots show the win rates (see \cref{sec:combined}) in \% when all seven methods are taken into account.}
\label{fig:combined_summary}
\end{figure*}

\section{Experiments}
\label{sec:combined}
We evaluate \methodname on two primary problem classes: unseen bioactivity tasks from Polaris (\cref{sec:polaris}) and activity-cliff benchmarks from MoleculeACE (\cref{sec:moleculeace}).

Our evaluation compares seven methods: MolFormer, CheMeleon, MiniMol, and \methodname, as well as novel TabPFN variants of CheMeleon and MiniMol, and GBM, a classical baseline that uses no pretraining: count-based ECFP4 fingerprints with RDKit 2D descriptors, given to a LightGBM tuned per task by randomised search cross-validated on the training split. This is the classical QSAR baseline that remains competitive on the leaderboards, and \citet{praski2025benchmarking} find that nearly all pretrained embedding models fail to improve on plain ECFP fingerprints.
We do not consider MolE \citep{mendezlucio2024mole}, Chemprop \citep{yang2019chemprop} or MolCLR \citep{wang2022molecular} in our comparisons because the first was shown to be weaker than MiniMol, while the latter two were shown to be weaker than CheMeleon in their respective publications.

Following the evaluation protocol of \citet{walters2024polaris} and \citet{burns2025chemeleon},
we incorporate model variability into the evaluation in order to allow statistically significant evaluation of the models' relative performance, even under multiple sources of variance.
This protocol performs simultaneous pairwise comparisons of all models, where a model is considered a ``winner'' on a specific task if its performance is statistically indistinguishable from the best-performing model (significance level $\alpha=0.05$, used for all statistical tests across the paper).
We aggregate these results to report the total win count and win rate across all tasks.
For pairwise comparisons we adopt the Tukey Honestly Significant Difference (HSD) test, as recommended by \citet{walters2024polaris}, and apply the Benjamini--Hochberg (BH) procedure~\citep{benjamini1995controlling} to control the False Discovery Rate (FDR) at $\alpha=0.05$ independently for each benchmark.
The same Tukey HSD + BH procedure is used for every pairwise comparison in this paper, including the ablation studies.
\cref{fig:combined_summary} summarizes the pairwise win counts across both benchmarks.

Additionally we compile a large ``leaderboard''-like table for 28 tasks in the Polaris benchmark, where we rank solely based on the mean result, ignoring variance (see \cref{tab:leaderboard}).
Although less statistically sound, this may be of interest as an alternative assessment with a wider range of methods. 

\subsection{Generalization to unseen biological tasks}
\label{sec:polaris}
To evaluate whether the model can generalize from pre-training to unseen biological contexts, we employ a subset of 28 bioactivity tasks from Polaris \cite{polaris2025,Huang2021tdc}.
These tasks cover ADMET and bioactivity properties, which lets us estimate the model's ability to extrapolate to novel chemical spaces and biological tasks, particularly in the low-data regime.

\methodname is the best method, or is statistically indistinguishable from the best method, on 100\% of Polaris tasks (\cref{tab:combined-results}) and achieves the lowest mean rank (3.71 over the full leaderboard, see \cref{tab:leaderboard}).
\begin{wraptable}{r}{0.6\textwidth}
\centering
\caption{Model performance on Polaris and MoleculeACE.
Error bars report the mean standard deviation across seeds.
}
\label{tab:combined-results}
\resizebox{\linewidth}{!}{%
\begin{tabular}{lcccccc}
\toprule
& \multicolumn{2}{c}{\textbf{Polaris}} & \multicolumn{3}{c}{\textbf{MoleculeACE}} \\
\cmidrule(lr){2-3} \cmidrule(lr){4-6}
\textbf{Model} & \textbf{Win \%} & \textbf{Rank $\downarrow$} & \textbf{Win \%} & \textbf{RMSE $\downarrow$} & \textbf{Cliff RMSE $\downarrow$} \\
\midrule
MolFormer                           & 10.7 & 17.57 & \phantom{0}0.0 & 0.748 {\scriptsize $\pm$ 0.018} & 0.854 {\scriptsize $\pm$ 0.027} \\
CheMeleon                           & 46.4 & 13.25 & 36.7 & 0.662 {\scriptsize $\pm$ 0.016} & 0.760 {\scriptsize $\pm$ 0.026} \\
MiniMol                             & 42.9 & 11.39 & 10.0 & 0.703 {\scriptsize $\pm$ 0.018} & 0.794 {\scriptsize $\pm$ 0.031} \\
\chemeleonpp                        & 42.9 & 11.46 & \underline{80.0} & \underline{0.641 {\scriptsize $\pm$ 0.013}} & \underline{0.741 {\scriptsize $\pm$ 0.020}} \\
GBM                   & 46.4 & 10.04 & 16.7 & 0.680 {\scriptsize $\pm$ 0.020} & 0.769 {\scriptsize $\pm$ 0.023} \\
\minimolpp                          & \underline{75.0} & \phantom{0}\underline{5.82} & 70.0 & 0.642 {\scriptsize $\pm$ 0.004} & 0.760 {\scriptsize $\pm$ 0.008} \\
\methodname                         & \textbf{100.0} & \phantom{0}\textbf{3.71} & \textbf{86.7} & \textbf{0.629 {\scriptsize $\pm$ 0.004}} & \textbf{0.737 {\scriptsize $\pm$ 0.006}} \\
\bottomrule
\end{tabular}}
\vspace*{-6ex}
\end{wraptable}
In addition, \minimolpp, a method which is also new to this paper, shows strong performance.

\subsection{Sensitivity to activity cliffs}
\label{sec:moleculeace}
MoleculeACE \cite{vanTilborg2022moleculeace} specifically assesses the model's performance on activity cliffs---structurally similar molecules with divergent potencies.
The primary metrics are both the overall Root Mean Squared Error (RMSE) and the RMSE on
``cliff'' compounds (those marking a steep change in activity).
The dataset partitioning facilitates this by distributing molecular series across splits,
allowing for a targeted evaluation of whether a model can transcend simple determinants of activity.

\begin{figure}[b]
\vspace{-1ex}
\centering
\includegraphics[width=0.9\columnwidth]{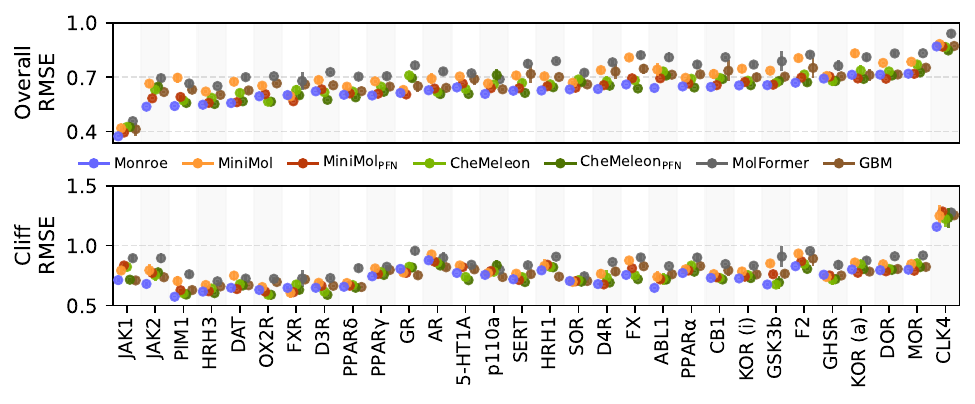}
\caption{Per-task Cliff RMSE on MoleculeACE for each method.}
\label{fig:moleculeace_boxplot}
\vspace{-1.5ex}
\end{figure}

On MoleculeACE \methodname leads on every metric: lowest overall RMSE (0.629), lowest Cliff RMSE (0.737), and the highest Win~\% (86.7\%, computed on overall RMSE). \chemeleonpp and \minimolpp are next, with the original baselines trailing further behind. \cref{fig:moleculeace_boxplot} shows the per-task Cliff RMSE for each method, while \cref{fig:per-task-moleculeace} gives the per-task pairwise comparisons.
The small gap to \minimolpp on Polaris widens substantially on this harder benchmark, suggesting that \methodname's structure-aware pretraining and stereochemistry rewiring may resolve local activity changes that fingerprint-only models miss.

\def\taskvariant#1{{\bf #1:}}

\subsection{Downstream adaptation for baseline models}
We seek to establish the most effective downstream adaptation strategy for each compared model on 28 Polaris benchmarks.
For each model, we compare three distinct strategies:

\textbf{Native}:
We use the adaptation method originally proposed for each model (typically a task-specific MLP head trained from scratch or finetuned with gradient descent).

\textbf{Pretrained embeddings ($^{\text{PT}}_{\text{PFN}}$):}
We use the frozen pretrained encoder to extract embeddings for TabPFN without any task-specific weight updates.

\textbf{Finetuned embeddings ($^{\text{FT}}_{\text{PFN}}$):}
We finetune the encoder on the downstream task before extracting embeddings for TabPFN. We hypothesize that while finetuning may adapt representations to the specific domain, it may also risk overfitting and potentially degrade the quality of signal for subsequent TabPFN-based inference.

\Cref{tab:downstream-adaptation} reports the within-model win rate for each strategy. The best strategy is model-dependent: MiniMol prefers frozen pretrained embeddings ($^{\text{PT}}_{\text{PFN}}$, 92.9\%), CheMeleon prefers finetuned embeddings ($^{\text{FT}}_{\text{PFN}}$, 96.4\%), and MolFormer prefers its native MLP head (92.9\%).

\begin{wraptable}{r}{0.33\textwidth}
\centering
\vspace{-1.3em}
\caption{Downstream adaptation comparison on Polaris. In this test we compare within variants of each method, not across methods.}
\label{tab:downstream-adaptation}
\small
\resizebox{\linewidth}{!}{%
\begin{tabular}{lccc}
\toprule
\textbf{Model} & \textbf{$^{\text{PT}}_{\text{PFN}}$} & \textbf{$^{\text{FT}}_{\text{PFN}}$} & \textbf{Native} \\
\midrule
MiniMol   & \textbf{92.9} & \underline{85.7} & 57.1 \\
CheMeleon & 75.0 & \textbf{96.4} & \underline{92.9} \\
MolFormer & 39.3 & \underline{71.4} & \textbf{92.9} \\
\bottomrule
\end{tabular}}
\end{wraptable}

For MiniMol, using pretrained embeddings with TabPFN substantially outperforms both finetuned embeddings and native adaptation; for CheMeleon, finetuned embeddings with TabPFN edge out the already-strong native adaptation. In both cases the TabPFN-based variant is the best for that model, and we carry the two forward as new methods, \minimolpp (MiniMol $^{\text{PT}}_{\text{PFN}}$) and \chemeleonpp (CheMeleon $^{\text{FT}}_{\text{PFN}}$), throughout the remainder of this paper, finding that they become strong additional baselines. MolFormer is the exception: its native MLP-head adaptation outperforms both PFN variants, so we keep MolFormer in its native form.

\subsection{Ablations}
\label{sec:ablations}

\begin{wraptable}{r}{0.55\textwidth}
\vspace{-1.35em}
\centering
\caption{Ablation study on Polaris and MoleculeACE. Pairwise testing is performed within each group separately.
Error bars report the mean standard deviation across 3 seeds.
}
\label{tab:ablations}
\resizebox{\linewidth}{!}{%
\begin{tabular}{lcccc}
\toprule
& \multicolumn{2}{c}{\textbf{Polaris}} & \multicolumn{2}{c}{\textbf{MoleculeACE}} \\
\cmidrule(lr){2-3} \cmidrule(lr){4-5}
\textbf{Variant} & \textbf{Win \%} & \textbf{Rank $\downarrow$} & \textbf{RMSE $\downarrow$} & \textbf{Cliff RMSE $\downarrow$} \\
\midrule
\multicolumn{5}{l}{\small{\emph{Multi-task balancing}}} \\
\midrule
\textbf{UW} & \textbf{75.0} & \textbf{3.71} & 0.629 {\scriptsize $\pm$ 0.003} & \textbf{0.737 {\scriptsize $\pm$ 0.005}} \\
EW & \underline{67.9} & \underline{3.71} & 0.633 {\scriptsize $\pm$ 0.003} & 0.743 {\scriptsize $\pm$ 0.006} \\
RLW & 42.9 & 3.96 & 0.634 {\scriptsize $\pm$ 0.004} & 0.746 {\scriptsize $\pm$ 0.006} \\
STCH & 39.3 & 4.32 & \textbf{0.627 {\scriptsize $\pm$ 0.003}} & \underline{0.738 {\scriptsize $\pm$ 0.006}} \\
DWA & 7.1 & 10.71 & 0.680 {\scriptsize $\pm$ 0.003} & 0.788 {\scriptsize $\pm$ 0.005} \\
\midrule
\multicolumn{5}{l}{\small{\emph{Conformer denoising}}} \\
\midrule
\textbf{$\checkmark$} & \textbf{92.9} & \textbf{3.71} & \textbf{0.629 {\scriptsize $\pm$ 0.003}} & \textbf{0.737 {\scriptsize $\pm$ 0.005}} \\
$\times$ & 60.7 & 4.61 & 0.638 {\scriptsize $\pm$ 0.003} & 0.749 {\scriptsize $\pm$ 0.005} \\
\midrule
\multicolumn{5}{l}{\small{\emph{Stereochemistry augmentation}}} \\
\midrule
\textbf{$\checkmark$} & \textbf{78.6} & \textbf{3.71} & \textbf{0.629 {\scriptsize $\pm$ 0.003}} & \textbf{0.737 {\scriptsize $\pm$ 0.005}} \\
$\times$ & 53.6 & 3.75 & 0.633 {\scriptsize $\pm$ 0.004} & 0.741 {\scriptsize $\pm$ 0.007} \\
\midrule
\multicolumn{5}{l}{\small{\emph{Decorrelation loss}}} \\
\midrule
\textbf{$\checkmark$ (w=8)} & \textbf{92.9} & \textbf{3.71} & 0.629 {\scriptsize $\pm$ 0.003} & \textbf{0.737 {\scriptsize $\pm$ 0.005}} \\
$\times$ (w=0) & 67.9 & 4.21 & \textbf{0.626 {\scriptsize $\pm$ 0.004}} & 0.739 {\scriptsize $\pm$ 0.005} \\
\midrule
\multicolumn{5}{l}{\small{\emph{Dataset}}} \\
\midrule
\textbf{All states + PCBA} & \textbf{89.3} & \textbf{3.71} & \textbf{0.629 {\scriptsize $\pm$ 0.003}} & \textbf{0.737 {\scriptsize $\pm$ 0.005}} \\
No PCBA & \underline{32.1} & \underline{5.36} & \underline{0.641 {\scriptsize $\pm$ 0.004}} & \underline{0.750 {\scriptsize $\pm$ 0.005}} \\
S0 only (no excited states) & 28.6 & 5.57 & 0.651 {\scriptsize $\pm$ 0.003} & 0.759 {\scriptsize $\pm$ 0.004} \\
\bottomrule
\end{tabular}}
\vspace{-2em}
\end{wraptable}

Table~\ref{tab:ablations} shows controlled ablation studies to isolate the contribution of individual components of \methodname. In each instance, alternatives were compared to the finally selected model, again under the Tukey HSD + BH pairwise evaluation protocol.

\textbf{Multi-task balancing.}
We compare five scalarization methods: EW, UW, DWA, RLW, and STCH. DWA is clearly harmful, whereas the rest land within noise of one another on both benchmarks. The UW is slightly better, whose per-task learnable variance down-weights noisy or already-saturated tasks, so we choose it for the final variant.

\textbf{Conformer denoising.}
We evaluate the contribution of the auxiliary coordinate denoising objective ($\mathcal{L}_{\mathrm{conf}}$) by comparing the full model against a variant trained without it.
Removing the objective collapses the Polaris win rate and degrades MoleculeACE on both overall and cliff RMSE.
We hypothesize that the 3D refinement signal forces the encoder to internalize geometric constraints that transfer to ADMET prediction.

\textbf{Stereochemistry augmentation.}
We assess the importance of our stereochemistry-aware graph rewiring by comparing it against a standard graph representation that ignores E/Z and R/S edges.
Its inclusion confers a clear advantage on the Polaris tasks, consistent with observations that many biological tasks are stereoselective, but is equivalent up to experimental variation on MoleculeACE.

\textbf{Decorrelation loss.}
We compare the full model with the embedding-decorrelation penalty against a variant without.
The penalty confers a decisive win on Polaris, but within-noise impact on MoleculeACE.
Our best hypothesis is that TabPFN on the classification tasks in Polaris is enhanced by decorrelation, while regression tasks in general are unaffected.

\textbf{Dataset.}
We assess the contribution of the two largest pretraining sources by training (i) on PM6 ground state only (excluding excited and charged-species tasks) and (ii) without the PCBA bioactivity assays. Removing either source degrades both Polaris win rate and MoleculeACE RMSE substantially, indicating that the chosen data sources are critical factor of the achieved performance.

\textbf{Downstream adaptation strategy.}
We compare three ways of adapting pretrained encoder to per-task predictions (\cref{tab:adaptation}). \emph{TabPFN} (the default): the encoder is held fixed and TabPFN performs in-context inference over the embeddings. \emph{MLP decoders}: a task-specific multi-layer perceptron is trained from scratch on the frozen embeddings (the MiniMol recipe). \emph{LoRA$\to$TabPFN}: low-rank adapters in the encoder's attention projections and pooling are finetuned per task through a linear head with a supervised loss; at inference, the head is discarded and TabPFN takes its place.

\begin{wraptable}{r}{0.5\textwidth}
\vspace{-3ex}
\centering
\caption{Downstream adaptation strategies on Monroe. Win~\% is computed pairwise within this group.}
\label{tab:adaptation}
\resizebox{\linewidth}{!}{%
\begin{tabular}{lcccc}
\toprule
& \multicolumn{2}{c}{\textbf{Polaris}} & \multicolumn{2}{c}{\textbf{MoleculeACE}} \\
\cmidrule(lr){2-3} \cmidrule(lr){4-5}
\textbf{Strategy} & \textbf{Win \%} & \textbf{Rank $\downarrow$} & \textbf{RMSE $\downarrow$} & \textbf{Cliff RMSE $\downarrow$} \\
\midrule
\textbf{TabPFN(v2)}    & \textbf{100.0} & \textbf{3.93} & \textbf{0.637 {\scriptsize $\pm$ 0.002}} & \textbf{0.740 {\scriptsize $\pm$ 0.003}} \\
MLP decoders           & 28.6           & 9.04          & 0.723 {\scriptsize $\pm$ 0.001}          & 0.806 {\scriptsize $\pm$ 0.007} \\
LoRA$\to$TabPFN        & \underline{89.3}           & \underline{4.93}          & \underline{0.653 {\scriptsize $\pm$ 0.001}}          & \underline{0.749 {\scriptsize $\pm$ 0.003}} \\
\bottomrule
\end{tabular}}
\vspace{-2.2em}
\end{wraptable}

Using TabPFN with a frozen encoder is a clear winner: adding per-task LoRA adapters on top of the same encoder does not improve over plain in-context inference, and a trained MLP head trails by a wide margin. We read this as evidence that in-context inference with TabPFN absorbs the role that downstream finetuning would otherwise play.

\subsection{Scaffold-split and test-partition evaluation}
\label{sec:scaffold}

\begin{wraptable}{r}{0.57\textwidth}
\vspace{-1.2em}
\centering
\caption{Polaris under repeated Bemis--Murcko scaffold splits grouped by metric. $1-\text{MAE}/\sigma_y$ is used for meaningful aggregation across scales. Scores are not comparable to \cref{tab:combined-results}, which uses the official benchmark split.}
\label{tab:scaffold}
\small
\resizebox{\linewidth}{!}{%

\begin{tabular}{lcccc}
\toprule
\textbf{Metric} & \textbf{$n$ tasks} & \methodname & \minimolpp & \chemeleonpp \\
\midrule
AUROC                   & 8 & \textbf{0.853 {\scriptsize $\pm$ 0.034}} & \underline{0.847 {\scriptsize $\pm$ 0.032}} & 0.811 {\scriptsize $\pm$ 0.049} \\
AUPRC                   & 5 & \textbf{0.706 {\scriptsize $\pm$ 0.035}} & \underline{0.701 {\scriptsize $\pm$ 0.045}} & 0.625 {\scriptsize $\pm$ 0.048} \\
$1-\text{MAE}/\sigma_y$ & 5 & \textbf{0.610 {\scriptsize $\pm$ 0.029}} & \underline{0.583 {\scriptsize $\pm$ 0.032}} & 0.547 {\scriptsize $\pm$ 0.038} \\
Pearson $r$             & 4 & \textbf{0.762 {\scriptsize $\pm$ 0.022}} & \underline{0.741 {\scriptsize $\pm$ 0.019}} & 0.642 {\scriptsize $\pm$ 0.026} \\
Spearman $r$            & 4 & \textbf{0.579 {\scriptsize $\pm$ 0.064}} & \underline{0.558 {\scriptsize $\pm$ 0.079}} & 0.541 {\scriptsize $\pm$ 0.078} \\
\bottomrule
\end{tabular}}
\end{wraptable}

We test sensitivity of the methods to test-partitioning and generalization to unseen scaffolds. Polaris withholds its test labels, so the split is constructed inside the official training split: molecules are grouped by Bemis--Murcko scaffold and whole groups are assigned to one side, at a fixed 80/20 ratio with five independent repeats per task, with every method evaluated on identical partitions.
Two of the 28 tasks (Fang RPPB and Fang HPPB) are excluded because splitting makes their test folds too small to measure. Roughly two thirds of Polaris molecules carry a unique scaffold, so the grouping constrains the assignment less than it might elsewhere. The achieved shift, as the mean nearest-neighbour Tanimoto from a test molecule to the training set, falls from 0.532 under random assignment to 0.443 under scaffold assignment.

\methodname obtains the best mean score in every metric group (\cref{tab:scaffold}), and under the Tukey HSD + BH protocol it is the best method, or statistically indistinguishable from it, on all 26 tasks.

\subsection{Downstream data regimes}
\label{sec:data-regime}

We also vary how much labelled data is given as support for in-context inference.  We subsample the support set to 1, 5, 10, 25 and 50\% of the labelled data, holding the test sets at full size. Subsampling is stratified for classification, with a floor of eight molecules.

\begin{table}[H]
\vspace{-1em}
\centering
\caption{Downstream performance as a function of labelled support-set size.}
\label{tab:data-regime}
\resizebox{\linewidth}{!}{%
\begin{tabular}{lccccccccc}
\toprule
& \multicolumn{3}{c}{\textbf{Polaris Rank $\downarrow$}} & \multicolumn{3}{c}{\textbf{MoleculeACE RMSE $\downarrow$}} & \multicolumn{3}{c}{\textbf{MoleculeACE Cliff RMSE $\downarrow$}} \\
\cmidrule(lr){2-4} \cmidrule(lr){5-7} \cmidrule(lr){8-10}
\textbf{Support} & \methodname & \minimolpp & \chemeleonpp & \methodname & \minimolpp & \chemeleonpp & \methodname & \minimolpp & \chemeleonpp \\
\midrule
1\%   & \textbf{23.29} & \underline{23.79} & 24.36 & \underline{1.173 {\scriptsize $\pm$ 0.019}} & \textbf{1.162 {\scriptsize $\pm$ 0.030}} & 1.247 {\scriptsize $\pm$ 0.021} & \underline{1.195 {\scriptsize $\pm$ 0.036}} & \textbf{1.181 {\scriptsize $\pm$ 0.043}} & 1.244 {\scriptsize $\pm$ 0.032} \\
5\%   & \textbf{20.64} & \underline{21.86} & 23.68 & \textbf{1.014 {\scriptsize $\pm$ 0.010}} & \underline{1.015 {\scriptsize $\pm$ 0.010}} & 1.116 {\scriptsize $\pm$ 0.006} & \textbf{1.048 {\scriptsize $\pm$ 0.016}} & \underline{1.050 {\scriptsize $\pm$ 0.016}} & 1.140 {\scriptsize $\pm$ 0.011} \\
10\%  & \textbf{17.54} & \underline{19.39} & 22.96 & \textbf{0.914 {\scriptsize $\pm$ 0.008}} & \underline{0.920 {\scriptsize $\pm$ 0.006}} & 1.022 {\scriptsize $\pm$ 0.009} & \textbf{0.968 {\scriptsize $\pm$ 0.013}} & \underline{0.977 {\scriptsize $\pm$ 0.007}} & 1.052 {\scriptsize $\pm$ 0.012} \\
25\%  & \textbf{11.93} & \underline{16.29} & 21.57 & \textbf{0.786 {\scriptsize $\pm$ 0.008}} & \underline{0.794 {\scriptsize $\pm$ 0.006}} & 0.854 {\scriptsize $\pm$ 0.010} & \textbf{0.867 {\scriptsize $\pm$ 0.014}} & \underline{0.880 {\scriptsize $\pm$ 0.010}} & 0.924 {\scriptsize $\pm$ 0.009} \\
50\%  & \phantom{0}\textbf{7.29} & \underline{10.57} & 18.50 & \textbf{0.707 {\scriptsize $\pm$ 0.005}} & \underline{0.717 {\scriptsize $\pm$ 0.004}} & 0.747 {\scriptsize $\pm$ 0.004} & \textbf{0.809 {\scriptsize $\pm$ 0.008}} & \underline{0.822 {\scriptsize $\pm$ 0.006}} & 0.851 {\scriptsize $\pm$ 0.004} \\
100\% & \phantom{0}\textbf{3.71} & \phantom{0}\underline{5.82} & 11.46 & \textbf{0.629 {\scriptsize $\pm$ 0.001}} & 0.642 {\scriptsize $\pm$ 0.000} & \underline{0.641 {\scriptsize $\pm$ 0.002}} & \textbf{0.737 {\scriptsize $\pm$ 0.001}} & 0.760 {\scriptsize $\pm$ 0.001} & \underline{0.741 {\scriptsize $\pm$ 0.004}} \\
\bottomrule
\end{tabular}}
\vspace{-1.5em}
\end{table}

\methodname has the lowest Polaris mean rank at every regime (\cref{tab:data-regime}).
On MoleculeACE it leads on both RMSE and cliff RMSE from 10\% upwards, while at 1 and 5\% the methods fall within each other's error bars, i.e. below roughly a hundred labelled molecules per task the support set is too small to separate the encoders.

\subsection{Model ensembling}
\label{sec:ensembling}

\begin{wraptable}{r}{0.55\textwidth}
\centering
\vspace{-1.35em}
\caption{Prediction-level ensembling of \methodname{}. \methodnameens{k} pools TabPFN predictions from the $k$ strongest models. \minimolpp{} added at $k{=}9$ and \chemeleonpp{} added at $k{=}10$.}
\label{tab:ensemble}
\resizebox{\linewidth}{!}{%
\begin{tabular}{lccc}
\toprule
\textbf{Method} & \textbf{Polaris Rank $\downarrow$} & \textbf{MACE RMSE $\downarrow$} & \textbf{MACE Cliff RMSE $\downarrow$} \\
\midrule
\methodname                                  & 3.71          & 0.629          & 0.737 \\
\midrule
\methodnameens{2}   & 3.18          & 0.623          & 0.733 \\
\methodnameens{4}   & \underline{2.86} & 0.620          & \underline{0.730} \\
\methodnameens{8}   & 3.00          & 0.621          & 0.732 \\
\methodnameens{9}   & 2.71          & \underline{0.619} & 0.732 \\
\methodnameens{10}  & \textbf{2.68} & \textbf{0.615} & \textbf{0.726} \\
\midrule
\minimolpp                                   & 5.82          & 0.642          & 0.760 \\
\chemeleonpp                                 & 11.46         & 0.641          & 0.741 \\
\bottomrule
\end{tabular}}
\vspace{-1 em}
\end{wraptable}

The top-ranked entries on the Polaris leaderboard are ensembles rather than single models~\citep{sypetkowski2024molgps}, hence we ask how far ensembling carries \methodname. We pool the per-molecule TabPFN predictions of several models by averaging regressions and class probabilities. We ensemble only models introduced in this paper: members are added by individual Polaris rank, first the variants from the ablation study (k=$\{2,4,8\}$, excluding DWA), then our two PFN-augmented baselines \minimolpp (k=$9$) and \chemeleonpp (k=$10$), where k is the number of models in the ensemble.

Table~\ref{tab:ensemble} reports the result. Pooling a few \methodname variants lowers the mean rank from 3.71 to 2.86 at k=$4$, after which adding weaker variants gives nothing back: k=$8$ is no better than k=$4$. What helps again are the two PFN-augmented baselines, individually much weaker than \methodname{} but wrong in different places, and the best rank (2.68) and the best activity-cliff RMSE (0.726) are both reached at k=$10$. The gain is therefore driven by error diversity rather than member quality.

\section{Discussion}
\label{sec:discussion}

This paper has introduced a new model for molecular property prediction, with fully open-source training and inference code~(see App.~\ref{app:code}). While it combines several individual innovations rather than a single ``big idea", we contend that it is useful in practice, and represents considerable step forward in the incorporation of prior-data-fitted models into the MFM pipeline.

The evaluation follows best current practices, but it is certainly worth noting that several flaws are known~\citep{walters2023better} in widely used benchmarks such as TDC.
These include inconsistent experimental conditions arising from aggregating data across multiple labs, unrealistic dynamic ranges, and varying classification cutoffs.
Datasets often suffer from curation errors, such as duplicates with conflicting labels, or contain high rates of assay artifacts.
Specific to TDC, we observe instances of chemically incorrect or irrelevant structures, stereochemical ambiguity in SMILES strings, and inconsistent handling of salts.
These issues are non-trivial to address and persist even in aggregated benchmarks. Polaris exposes subsets of TDC, making these criticisms directly relevant to parts of our evaluation.

Conformer denoising helps downstream accuracy, and removing it hurts both benchmarks (\cref{sec:ablations}). Still, \methodname is not a good conformation predictor, and under our training setup its predicted geometries are not better than the input conformers. We don't consider this a shortcoming, since structure determination is not a goal of the model. It arises from the task weighting: equal weighting does learn to denoise, but uncertainty weighting, which we use because it is far stronger downstream, lets the property and bioassay tasks take the priority over denoising.

Separately, \methodname has the lowest RMSE of any method we evaluate on MoleculeACE, on both the full test set and the activity-cliff subset (\cref{tab:combined-results}). Yet the cliff error still sits well above the overall error, as it does for every method, indicating that the benchmark is not solved. Predicting the sharp, local changes in activity that define cliffs remains an open problem.

\section{Societal impact}
\label{sec:impact}

This work develops machine learning methods for learning molecular representations that can be applied to scientific and industrial tasks such as molecular property prediction and early-stage drug discovery.
By improving representation quality, evaluation rigor, and downstream adaptation, the methods introduced here may reduce the cost and time required for exploratory molecular design.

As with many general-purpose modeling tools, these methods could also be applied in other domains involving molecular data, with impacts that depend on the specific context of use. The work does not introduce new biological agents, chemical processes, or experimental protocols, and its potential societal effects arise primarily through downstream applications chosen by practitioners.

\begin{ack}
We thank Emmanuel Noutahi for his feedback on the manuscript and Dexiong Chen for useful discussions.
\end{ack}

\bibliography{references}

@article{kabsch1976solution,
  title     = {A solution for the best rotation to relate two sets of vectors},
  author    = {Kabsch, Wolfgang},
  journal   = {Acta Crystallographica Section A},
  year      = {1976},
  volume    = {32},
  number    = {5},
  pages     = {922--923},
  doi       = {10.1107/S0567739476001873},
  url       = {https://doi.org/10.1107/S0567739476001873},
  publisher = {International Union of Crystallography}
}

@article{jumper2021alphafold,
	author = {Jumper, John and Evans, Richard and Pritzel, Alexander and Green, Tim and Figurnov, Michael and Ronneberger, Olaf and Tunyasuvunakool, Kathryn and Bates, Russ and Žídek, Augustin and Potapenko, Anna and Bridgland, Alex and Meyer, Clemens and Kohl, Simon A. A. and Ballard, Andrew J. and Cowie, Andrew and Romera-Paredes, Bernardino and Nikolov, Stanislav and Jain, Rishub and Adler, Jonas and Back, Trevor and Petersen, Stig and Reiman, David and Clancy, Ellen and Zielinski, Michal and Steinegger, Martin and Pacholska, Michalina and Berghammer, Tamas and Bodenstein, Sebastian and Silver, David and Vinyals, Oriol and Senior, Andrew W. and Kavukcuoglu, Koray and Kohli, Pushmeet and Hassabis, Demis},
	journal = {Nature},
	month = {7},
	number = {7873},
	pages = {583--589},
	title = {{Highly accurate protein structure prediction with AlphaFold}},
	volume = {596},
	year = {2021},
	doi = {10.1038/s41586-021-03819-2},
	url = {https://doi.org/10.1038/s41586-021-03819-2},
}

@misc{dwivedi2022graph,
  title     = {Graph Neural Networks with Learnable Structural and Positional Representations},
  author    = {Dwivedi, Vijay Prakash and Luu, Anh Tuan and Laurent, Thomas and Bengio, Yoshua and Bresson, Xavier},
  booktitle = {International Conference on Learning Representations (ICLR)},
  year      = {2022},
  url       = {https://openreview.net/forum?id=wTTjnvGphYj}
}

@article{ying2018hierarchical,
	author = {Ying, Zhitao and You, Jiaxuan and Morris, Christopher and Ren, Xiang and Hamilton, William L. and Leskovec, Jure},
	journal = {Neural Information Processing Systems},
	month = {12},
	pages = {4805--4815},
	title = {{Hierarchical graph representation learning with differentiable pooling}},
	volume = {31},
	year = {2018},
	url = {http://papers.nips.cc/paper/7729-hierarchical-graph-representation-learning-with-differentiable-pooling.pdf},
}

@misc{gilmer2017neural,
	author = {Gilmer, Justin and Schoenholz, Samuel S. and Riley, Patrick F. and Vinyals, Oriol and Dahl, George E.},
	month = {7},
	title = {{Neural message passing for quantum chemistry}},
	year = {2017},
	url = {http://proceedings.mlr.press/v70/gilmer17a.html},
}

@misc{li2016gated,
	author = {Li, Yujia and Tarlow, Daniel and Brockschmidt, Marc and Zemel, Richard},
	month = {11},
	title = {{Gated Graph sequence neural networks}},
	year = {2015},
	url = {https://arxiv.org/abs/1511.05493},
}

@misc{lee2019set,
	author = {Lee, Juho and Lee, Yoonho and Kim, Jungtaek and Kosiorek, Adam R. and Choi, Seungjin and Teh, Yee Whye},
	month = {10},
	title = {{Set Transformer: A framework for Attention-based Permutation-Invariant Neural Networks}},
	year = {2018},
	url = {https://arxiv.org/abs/1810.00825},
}

@article{lu2023unimolplus,
	author = {Lu, Shuqi and Gao, Zhifeng and He, Di and Zhang, Linfeng and Ke, Guolin},
	journal = {arXiv (Cornell University)},
	month = {3},
	title = {{Highly Accurate Quantum Chemical Property Prediction with Uni-Mol+}},
	year = {2023},
	doi = {10.48550/arxiv.2303.16982},
	url = {http://arxiv.org/abs/2303.16982},
}

@misc{nikitin2025loqi,
	author = {Nikitin, Filipp and Anstine, Dylan M. and Zubatyuk, Roman and Paliwal, Saee Gopal and Isayev, Olexandr},
	journal = {ChemRxiv},
	month = {8},
	title = {{Scalable Low-Energy Molecular Conformer Generation with Quantum Mechanical Accuracy}},
	year = {2025},
	doi = {10.26434/chemrxiv-2025-k4h7v},
	url = {https://doi.org/10.26434/chemrxiv-2025-k4h7v},
}

@article{klaser2024minimol,
	author = {Kläser, Kerstin and Banaszewski, Błażej and Maddrell-Mander, Samuel and McLean, Callum and Müller, Luis and Parviz, Ali and Huang, Shenyang and Fitzgibbon, Andrew},
	month = {4},
	title = {{$\texttt{MiniMol}$: A Parameter-Efficient Foundation Model for Molecular Learning}},
	year = {2024},
	doi = {10.48550/arxiv.2404.14986},
	url = {http://arxiv.org/abs/2404.14986},
}

@misc{sypetkowski2024molgps,
  title         = {On the Scalability of {GNN}s for Molecular Graphs},
  author        = {Sypetkowski, Maciej and Wenkel, Frederik and Poursafaei, Farimah and Dickson, Nia and Suri, Karush and Fradkin, Philip and Beaini, Dominique},
  year          = {2024},
  eprint        = {2404.11568},
  archivePrefix = {arXiv},
  primaryClass  = {cs.LG},
  doi           = {10.48550/arXiv.2404.11568},
  url           = {https://arxiv.org/abs/2404.11568}
}

@article{hollmann2025tabpfn,
  title   = {Accurate predictions on small data with a tabular foundation model},
  author  = {Hollmann, Noah and M{\"u}ller, Samuel and Purucker, Lennart and Krishnakumar, Arjun and K{\"o}rfer, Max and Hoo, Shi Bin and Schirrmeister, Robin Tibor and Hutter, Frank},
  journal = {Nature},
  year    = {2025},
  volume  = {637},
  number  = {8045},
  pages   = {319--326},
  doi     = {10.1038/s41586-024-08328-6},
  url     = {https://www.nature.com/articles/s41586-024-08328-6}
}

@article{mendezlucio2024mole,
  title   = {{MolE}: a foundation model for molecular graphs using disentangled attention},
  author  = {M{\'e}ndez-Lucio, Oscar and Nicolaou, Christos A. and Earnshaw, Berton},
  journal = {Nature Communications},
  year    = {2024},
  volume  = {15},
  number  = {1},
  pages   = {9431},
  doi     = {10.1038/s41467-024-53751-y},
  url     = {https://www.nature.com/articles/s41467-024-53751-y}
}

@inproceedings{pmlr-v202-ma23c,
  title     = {Graph Inductive Biases in Transformers without Message Passing},
  author    = {Ma, Liheng and Lin, Chen and Lim, Derek and Romero-Soriano, Adriana and Dokania, Puneet K. and Coates, Mark and Torr, Philip and Lim, Ser-Nam},
  booktitle = {Proceedings of the 40th International Conference on Machine Learning},
  pages     = {23321--23337},
  year      = {2023},
  editor    = {Krause, Andreas and Brunskill, Emma and Cho, Kyunghyun and Engelhardt, Barbara and Sabato, Sivan and Scarlett, Jonathan},
  volume    = {202},
  series    = {Proceedings of Machine Learning Research},
  month     = {23--29 Jul},
  publisher = {PMLR},
  url       = {https://proceedings.mlr.press/v202/ma23c.html}
}

@article{yang2019chemprop,
  title   = {Analyzing Learned Molecular Representations for Property Prediction},
  author  = {Yang, Kevin and Swanson, Kyle and Jin, Wengong and Coley, Connor and Eiden, Philipp and Gao, Hua and Guzman-Perez, Angel and Hopper, Timothy and Kelley, Brian and Mathea, Miriam and Palmer, Andrew and Settels, Volker and Jaakkola, Tommi and Jensen, Klavs and Barzilay, Regina},
  journal = {Journal of Chemical Information and Modeling},
  year    = {2019},
  volume  = {59},
  number  = {8},
  pages   = {3370--3388},
  doi     = {10.1021/acs.jcim.9b00237},
  url     = {https://pubs.acs.org/doi/10.1021/acs.jcim.9b00237}
}

@InProceedings{lin2022smooth,
  title = 	 {Smooth Tchebycheff Scalarization for Multi-Objective Optimization},
  author =       {Lin, Xi and Zhang, Xiaoyuan and Yang, Zhiyuan and Liu, Fei and Wang, Zhenkun and Zhang, Qingfu},
  booktitle = 	 {Proceedings of the 41st International Conference on Machine Learning},
  pages = 	 {30479--30509},
  year = 	 {2024},
  editor = 	 {Salakhutdinov, Ruslan and Kolter, Zico and Heller, Katherine and Weller, Adrian and Oliver, Nuria and Scarlett, Jonathan and Berkenkamp, Felix},
  volume = 	 {235},
  series = 	 {Proceedings of Machine Learning Research},
  month = 	 {21--27 Jul},
  publisher =    {PMLR},
  url = 	 {https://proceedings.mlr.press/v235/lin24y.html}
}

@article{vanTilborg2022moleculeace,
  title   = {Exposing the Limitations of Molecular Machine Learning with Activity Cliffs},
  author  = {van Tilborg, Derek and Alenicheva, Alisa and Grisoni, Francesca},
  journal = {Journal of Chemical Information and Modeling},
  year    = {2022},
  volume  = {62},
  number  = {23},
  pages   = {5938--5951},
  doi     = {10.1021/acs.jcim.2c01073},
  url     = {https://pubs.acs.org/doi/10.1021/acs.jcim.2c01073}
}

@article{walters2024polaris,
  title   = {Practically Significant Method Comparison Protocols for Machine Learning in Small Molecule Drug Discovery},
  author  = {Ash, Jeremy R. and Wognum, Cas and Rodr\'iguez-P\'erez, Raquel and Aldeghi, Matteo and Cheng, Alan C. and Clevert, Djork-Arn\'e and Engkvist, Ola and Fang, Cheng and Price, Daniel J. and Hughes-Oliver, Jacqueline M. and Walters, W. Patrick},
  journal = {Journal of Chemical Information and Modeling},
  year    = {2025},
  volume  = {65},
  number  = {18},
  pages   = {9398--9411},
  doi     = {10.1021/acs.jcim.5c01609},
  url     = {https://pubs.acs.org/doi/10.1021/acs.jcim.5c01609}
}

@misc{burns2025chemeleon,
  title         = {Descriptor-based Foundation Models for Molecular Property Prediction},
  author        = {Burns, Jackson W. and Zalte, Akshat S. and Green, William H.},
  year          = {2025},
  eprint        = {2506.15792},
  archivePrefix = {arXiv},
  primaryClass  = {cs.LG},
  doi           = {10.48550/arXiv.2506.15792},
  url           = {https://arxiv.org/abs/2506.15792},
  note          = {Introduces {CheMeleon}}
}

@article{Nakata2020PubChemQC,
  title   = {{PubChemQC} {PM6}: Data Sets of 221 Million Molecules with Optimized Molecular Geometries and Electronic Properties},

  author  = {Nakata, Maho and Shimazaki, Tomomi and Hashimoto, Masatomo and Maeda, Toshiyuki},
  journal = {Journal of Chemical Information and Modeling},
  year    = {2020},
  volume  = {60},
  number  = {12},
  pages   = {5891--5899},
  doi     = {10.1021/acs.jcim.0c00740},
  url     = {https://pubs.acs.org/doi/10.1021/acs.jcim.0c00740}
}

@inproceedings{Beaini2024Towards,
  title     = {Towards Foundational Models for Molecular Learning on Large-Scale Multi-Task Datasets},
  author    = {Beaini, Dominique and Huang, Shenyang and Cunha, Joao Alex and Li, Zhiyi and Moisescu-Pareja, Gabriela and Dymov, Oleksandr and Maddrell-Mander, Samuel and McLean, Callum and Wenkel, Frederik and M{\"u}ller, Luis and Mohamud, Jama Hussein and Parviz, Ali and Craig, Michael and Koziarski, Micha{\l} and Lu, Jiarui and Zhu, Zhaocheng and Gabellini, Cristian and Kl{\"a}ser, Kerstin and Dean, Josef and Wognum, Cas and Sypetkowski, Maciej and Rabusseau, Guillaume and Rabbany, Reihaneh and Tang, Jian and Morris, Christopher and Koutis, Ioannis and Ravanelli, Mirco and Wolf, Guy and Tossou, Prudencio and Mary, Hadrien and Bois, Therence and Fitzgibbon, Andrew W. and Banaszewski, B{\l}a{\.z}ej and Martin, Chad and Masters, Dominic},

  booktitle = {The Twelfth International Conference on Learning Representations},
  year      = {2024},
  url       = {https://openreview.net/forum?id=Zc2aIcucwc}
}

@article{stewart2007optimization,
  title     = {Optimization of parameters for semiempirical methods {V}: modification of {NDDO} approximations and application to 70 elements},

  author    = {Stewart, James J. P.},
  journal   = {Journal of Molecular Modeling},
  year      = {2007},
  volume    = {13},
  number    = {12},
  pages     = {1173--1213},
  doi       = {10.1007/s00894-007-0233-4},
  url       = {https://link.springer.com/article/10.1007/s00894-007-0233-4},
  publisher = {Springer}
}

@article{ross2022molformer,
  title     = {Large-scale chemical language representations capture molecular structure and properties},
  author    = {Ross, Jerret and Belgodere, Brian and Chenthamarakshan, Vijil and Padhi, Inkit and Mroueh, Youssef and Das, Payel},
  journal   = {Nature Machine Intelligence},
  year      = {2022},
  volume    = {4},
  number    = {12},
  pages     = {1256--1264},
  doi       = {10.1038/s42256-022-00580-7},
  url       = {https://www.nature.com/articles/s42256-022-00580-7},
  publisher = {Nature Publishing Group}
}

@inproceedings{ying2021do,
  title     = {Do Transformers Really Perform Bad for Graph Representation?},
  author    = {Ying, Chengxuan and Cai, Tianle and Luo, Shengjie and Zheng, Shuxin and Ke, Guolin and He, Di and Shen, Yanming and Liu, Tie-Yan},
  booktitle = {Advances in Neural Information Processing Systems},
  volume    = {34},
  pages     = {28877--28888},
  year      = {2021},
  url       = {https://proceedings.neurips.cc/paper/2021/file/f1c1592588411002af340cbaedd6fc33-Paper.pdf}
}

@misc{walters2023better,
  title  = {We Need Better Benchmarks for Machine Learning in Drug Discovery},
  author = {Walters, Pat},
  year   = {2023},
  month  = aug,
  note   = {Blog post, Practical Cheminformatics},
  url    = {https://practicalcheminformatics.blogspot.com/2023/08/we-need-better-benchmarks-for-machine.html}
}

@inproceedings{Huang2021tdc,
  title     = {Therapeutics Data Commons: Machine Learning Datasets and Tasks for Drug Discovery and Development},
  author    = {Huang, Kexin and Fu, Tianfan and Gao, Wenhao and Zhao, Yue and Roohani, Yusuf and Leskovec, Jure and Coley, Connor W. and Xiao, Cao and Sun, Jimeng and Zitnik, Marinka},
  booktitle = {Advances in Neural Information Processing Systems (NeurIPS) Datasets and Benchmarks Track},
  year      = {2021}
}

@article{moriwaki2018mordred,
  title   = {Mordred: a molecular descriptor calculator},
  author  = {Moriwaki, Hirotomo and Tian, Yu-Shi and Kawashita, Naoaki and Takagi, Toshihisa},
  journal = {Journal of Cheminformatics},
  year    = {2018},
  volume  = {10},
  number  = {1},
  pages   = {4},
  doi     = {10.1186/s13321-018-0258-y},
}

@article{trannguyen2020litpcba,
  title   = {{LIT-PCBA}: An unbiased data set for machine learning and virtual screening},
  author  = {Tran-Nguyen, Viet-Khoa and Célien Jacquemard and Rognan, Didier},
  journal = {Journal of Chemical Information and Modeling},
  year    = {2020},
  volume  = {60},
  number  = {9},
  pages   = {4263--4273},
  doi     = {10.1021/acs.jcim.0c00155},
}

@software{polaris2025,
  title        = {polaris-hub/polaris: Version 0.13.0},
  author       = {Wognum, Cas and Zhu, Lu and Mary, Hadrien and St-Laurent, Julien and {Larissa} and Quirke, Andrew and Hounwanou, Honor{\'e} and {roselyn!} and Li, Jack and Zhu, Kun and Whitfield, Shawn and {Daniel} and Burns, Jackson and Howe, Kira and {Howe (McLean)}, Kira and {felix}},
  year         = {2025},
  version      = {0.13.0},
  publisher    = {Zenodo},
  doi          = {10.5281/zenodo.15610218},
  url          = {https://doi.org/10.5281/zenodo.15610218}
}

@inproceedings{kendall2018uncertainty,
  title     = {Multi-Task Learning Using Uncertainty to Weigh Losses for Scene Geometry and Semantics},
  author    = {Kendall, Alex and Gal, Yarin and Cipolla, Roberto},
  booktitle = {Proceedings of the IEEE Conference on Computer Vision and Pattern Recognition (CVPR)},
  pages     = {7482--7491},
  year      = {2018},
  url       = {https://openaccess.thecvf.com/content_cvpr_2018/papers/Kendall_Multi-Task_Learning_Using_CVPR_2018_paper.pdf}
}

@inproceedings{liu2019endtoend,
  title     = {End-to-End Multi-Task Learning with Attention},
  author    = {Liu, Shikun and Johns, Edward and Davison, Andrew J.},
  booktitle = {Proceedings of the IEEE/CVF Conference on Computer Vision and Pattern Recognition (CVPR)},
  pages     = {1871--1880},
  year      = {2019},
  url       = {https://openaccess.thecvf.com/content_CVPR_2019/papers/Liu_End-To-End_Multi-Task_Learning_With_Attention_CVPR_2019_paper.pdf}
}

@article{lin2022rlw,
  title   = {Reasonable Effectiveness of Random Weighting: A Litmus Test for Multi-Task Learning},
  author  = {Lin, Baijiong and Ye, Feiyang and Zhang, Yu and Tsang, Ivor W.},
  journal = {Transactions on Machine Learning Research (TMLR)},
  year    = {2022},
  url     = {https://openreview.net/forum?id=jjtFD8A1Wx}
}

@article{lin2023libmtl,
  title={{LibMTL}: A {P}ython Library for Multi-Task Learning},
  author={Lin, Baijiong and Zhang, Yu},

  journal={Journal of Machine Learning Research},
  volume={24},
  number={209},
  pages={1--7},
  year={2023}
}

@article{wang2022molecular,
  title={Molecular contrastive learning of representations via graph neural networks},
  author={Wang, Yuyang and Wang, Jianren and Cao, Zhonglin and Barati Farimani, Amir},
  journal={Nature Machine Intelligence},
  volume={4},
  number={3},
  pages={279--287},
  year={2022},
  publisher={Nature Publishing Group},
  doi={10.1038/s42256-022-00447-x},
  url={https://doi.org/10.1038/s42256-022-00447-x}
}

@inproceedings{velivckovic2018gat,
  title     = {Graph Attention Networks},
  author    = {Veli{\v{c}}kovi{\'c}, Petar and Cucurull, Guillem and Casanova, Arantxa and Romero, Adriana and Li\`o, Pietro and Bengio, Yoshua},
  booktitle = {International Conference on Learning Representations (ICLR)},
  year      = {2018},
  url       = {https://arxiv.org/abs/1710.10903}
}

@misc{landrum2016rdkit,
  author = {Landrum, Greg},
  title = {{RDKit}: Open-source cheminformatics},
  year = {2016},
  url = {http://www.rdkit.org}
}

@article{benjamini1995controlling,
  title={Controlling the false discovery rate: a practical and powerful approach to multiple testing},
  author={Benjamini, Yoav and Hochberg, Yosef},
  journal={Journal of the Royal Statistical Society: Series B (Methodological)},
  volume={57},
  number={1},
  pages={289--300},
  year={1995},
  publisher={Wiley Online Library}
}

@inproceedings{bardes2022vicreg,
  title     = {{VICR}eg: Variance-Invariance-Covariance Regularization for Self-Supervised Learning},
  author    = {Bardes, Adrien and Ponce, Jean and LeCun, Yann},
  booktitle = {International Conference on Learning Representations (ICLR)},
  year      = {2022},
  url       = {https://openreview.net/forum?id=xm6YD62D1Ub}
}

@article{lin2017focal,
  title     = {Focal Loss for Dense Object Detection},
  author    = {Lin, Tsung-Yi and Goyal, Priya and Girshick, Ross and He, Kaiming and Doll{\'a}r, Piotr},
  journal   = {IEEE International Conference on Computer Vision (ICCV)},
  year      = {2017}
}

@misc{praski2025benchmarking,
  title         = {Benchmarking Pretrained Molecular Embedding Models For Molecular Representation Learning},
  author        = {Praski, Mateusz and Adamczyk, Jakub and Czech, Wojciech},
  year          = {2025},
  eprint        = {2508.06199},
  archivePrefix = {arXiv},
  primaryClass  = {cs.LG},
  doi           = {10.48550/arXiv.2508.06199},
  url           = {https://arxiv.org/abs/2508.06199}
}
\bibliographystyle{plainnat}

\newpage
\appendix

\section{Detailed results}
\label{sec:detailed-results}

\subsection{Benchmark details}

We compile the leaderboard in \Cref{tab:leaderboard} by joining the results from two separate resources: the Therapeutic Data Commons (TDC) ADMET leaderboard and the Polaris benchmarks. Some methods present on the Polaris boards have no reference to a codebase or paper; we filter those out and consider only methods referenced to an implementation. TDC provides standard deviation numbers across seeds, whereas Polaris only provides a single number per metric, without any error bars. For that reason, in order to consolidate them, we drop the standard deviations from the table and present only the mean results. We replace as-reported results with our own evaluations for \methodname, \minimolpp, \chemeleonpp, MiniMol, MolFormer, and CheMeleon where available.

\definecolor{colorMonroe}{HTML}{6666FF}
\definecolor{colorMiniMolPFN}{HTML}{BB3909}
\definecolor{colorMiniMol}{HTML}{FF9933}
\definecolor{colorCheMeleonPFN}{HTML}{4D7301}
\definecolor{colorCheMeleon}{HTML}{7BB601}
\definecolor{colorMolFormer}{HTML}{666666}

\begin{table*}[!ht]
\centering
\caption{Consolidated leaderboard from Polaris and TDC ADMET benchmarks. \textbf{\methodnameens{10}} (black) is the ensemble of \cref{tab:ensemble}. We replace as-reported results with our own evaluations for \textcolor{colorMonroe}{\textbf{\methodname}}, \textcolor{colorMiniMol}{\textbf{MiniMol}}, \textcolor{colorMiniMolPFN}{\textbf{\minimolpp}}, \textcolor{colorCheMeleon}{\textbf{CheMeleon}}, \textcolor{colorCheMeleonPFN}{\textbf{\chemeleonpp}} and \textcolor{colorMolFormer}{\textbf{MolFormer}}.}
\label{tab:leaderboard}
\setlength{\tabcolsep}{2pt}
{\footnotesize
\resizebox{\textwidth}{!}{%
\begin{tabular}{lrl ccccccccccccccc}
\toprule
\textbf{Dataset} & \textbf{Size} & \textbf{Metric} & \textbf{1} & \textbf{2} & \textbf{3} & \textbf{4} & \textbf{5} & \textbf{6} & \textbf{7} & \textbf{8} & \textbf{9} & \textbf{10} & \textbf{11} & \textbf{12} & \textbf{13} & \textbf{14} & \textbf{15} \\
\midrule
CYP2D6 Veith & 13,130 & AUPRC ($\uparrow$) & 0.801 & 0.790 & \textcolor{black}{\textbf{0.749}} & \textcolor{colorMonroe}{\textbf{0.739}} & 0.739 & 0.737 & \textcolor{colorMiniMolPFN}{\textbf{0.726}} & 0.723 & 0.721 & 0.718 & \textcolor{colorCheMeleonPFN}{\textbf{0.716}} & \textcolor{colorCheMeleon}{\textbf{0.705}} & \textcolor{colorMiniMol}{\textbf{0.698}} & 0.685 & 0.673 \\
CYP3A4 Veith & 12,328 & AUPRC ($\uparrow$) & 0.919 & 0.916 & 0.904 & 0.902 & \textcolor{black}{\textbf{0.901}} & \textcolor{colorMonroe}{\textbf{0.895}} & 0.895 & \textcolor{colorMiniMolPFN}{\textbf{0.884}} & 0.884 & 0.881 & 0.876 & 0.875 & \textcolor{colorCheMeleonPFN}{\textbf{0.874}} & \textcolor{colorCheMeleon}{\textbf{0.873}} & 0.867 \\
CYP2C9 Veith & 12,092 & AUPRC ($\uparrow$) & 0.895 & 0.859 & 0.839 & \textcolor{black}{\textbf{0.835}} & 0.833 & 0.829 & \textcolor{colorMonroe}{\textbf{0.826}} & \textcolor{colorMiniMolPFN}{\textbf{0.825}} & \textcolor{colorMiniMol}{\textbf{0.816}} & \textcolor{colorCheMeleonPFN}{\textbf{0.796}} & 0.789 & 0.786 & 0.783 & 0.777 & \textcolor{colorCheMeleon}{\textbf{0.776}} \\
Solubility AqSolDB & 9,982 & MAE ($\downarrow$) & \textcolor{black}{\textbf{0.657}} & \textcolor{colorMonroe}{\textbf{0.681}} & \textcolor{colorMiniMolPFN}{\textbf{0.719}} & 0.726 & 0.743 & \textcolor{colorMolFormer}{\textbf{0.744}} & 0.761 & 0.775 & 0.776 & 0.789 & 0.792 & 0.796 & \textcolor{colorMiniMol}{\textbf{0.796}} & 0.827 & 0.828 \\
LD50 Zhu & 5,907 & MAE ($\downarrow$) & \textcolor{colorCheMeleonPFN}{\textbf{0.533}} & \textcolor{black}{\textbf{0.547}} & 0.552 & \textcolor{colorMonroe}{\textbf{0.555}} & 0.573 & \textcolor{colorCheMeleon}{\textbf{0.576}} & 0.588 & \textcolor{colorMiniMolPFN}{\textbf{0.593}} & 0.606 & \textcolor{colorMiniMol}{\textbf{0.612}} & 0.614 & 0.614 & 0.621 & 0.622 & 0.625 \\
Ames & 5,821 & AUROC ($\uparrow$) & \textcolor{black}{\textbf{0.881}} & \textcolor{colorMonroe}{\textbf{0.875}} & 0.871 & 0.870 & 0.869 & 0.868 & 0.868 & \textcolor{colorCheMeleon}{\textbf{0.866}} & 0.866 & \textcolor{colorMiniMolPFN}{\textbf{0.865}} & \textcolor{colorCheMeleonPFN}{\textbf{0.862}} & 0.852 & 0.850 & \textcolor{colorMiniMol}{\textbf{0.847}} & 0.847 \\
Lipophilicity AZ & 3,360 & MAE ($\downarrow$) & 0.392 & \textcolor{colorMonroe}{\textbf{0.397}} & \textcolor{black}{\textbf{0.399}} & 0.411 & 0.424 & \textcolor{colorMiniMolPFN}{\textbf{0.444}} & \textcolor{colorCheMeleon}{\textbf{0.463}} & \textcolor{colorCheMeleonPFN}{\textbf{0.467}} & 0.467 & 0.470 & 0.479 & \textcolor{colorMiniMol}{\textbf{0.492}} & \textcolor{colorMolFormer}{\textbf{0.508}} & 0.512 & 0.515 \\
PPBR AZ & 2,231 & MAE ($\downarrow$) & \textcolor{black}{\textbf{6.589}} & \textcolor{colorMonroe}{\textbf{6.801}} & 7.004 & 7.026 & \textcolor{colorCheMeleon}{\textbf{7.098}} & \textcolor{colorMiniMolPFN}{\textbf{7.235}} & 7.440 & 7.526 & \textcolor{colorCheMeleonPFN}{\textbf{7.599}} & 7.660 & 7.788 & 7.914 & 7.990 & 8.200 & 8.288 \\
Fang HCLint & 2,229 & Pearson ($\uparrow$) & \textcolor{black}{\textbf{0.791}} & \textcolor{colorMonroe}{\textbf{0.784}} & 0.778 & \textcolor{colorMiniMolPFN}{\textbf{0.742}} & \textcolor{colorMiniMol}{\textbf{0.727}} & 0.720 & 0.717 & \textcolor{colorCheMeleonPFN}{\textbf{0.710}} & \textcolor{colorMolFormer}{\textbf{0.707}} & \textcolor{colorCheMeleon}{\textbf{0.695}} & 0.674 &  &  &  &  \\
Fang RClint & 2,218 & Pearson ($\uparrow$) & 0.798 & \textcolor{black}{\textbf{0.786}} & 0.784 & \textcolor{colorMonroe}{\textbf{0.775}} & \textcolor{colorMiniMolPFN}{\textbf{0.746}} & \textcolor{colorMiniMol}{\textbf{0.735}} & \textcolor{colorCheMeleonPFN}{\textbf{0.729}} & \textcolor{colorMolFormer}{\textbf{0.720}} & 0.714 & \textcolor{colorCheMeleon}{\textbf{0.709}} & 0.698 & 0.660 &  &  &  \\
Fang Perm & 1,919 & Pearson ($\uparrow$) & 0.879 & \textcolor{black}{\textbf{0.866}} & 0.860 & \textcolor{colorMonroe}{\textbf{0.851}} & \textcolor{colorMiniMolPFN}{\textbf{0.824}} & \textcolor{colorMiniMol}{\textbf{0.808}} & \textcolor{colorCheMeleonPFN}{\textbf{0.795}} & \textcolor{colorMolFormer}{\textbf{0.791}} & \textcolor{colorCheMeleon}{\textbf{0.773}} & 0.772 & 0.762 & 0.714 &  &  &  \\
BBB Martins & 1,624 & AUROC ($\uparrow$) & \textcolor{black}{\textbf{0.936}} & 0.933 & \textcolor{colorMonroe}{\textbf{0.932}} & 0.928 & \textcolor{colorMiniMol}{\textbf{0.922}} & 0.920 & \textcolor{colorMiniMolPFN}{\textbf{0.920}} & 0.916 & 0.915 & 0.913 & 0.912 & 0.910 & 0.908 & 0.906 & 0.905 \\
Fang Solu & 1,578 & Pearson ($\uparrow$) & 0.770 & 0.764 & \textcolor{black}{\textbf{0.734}} & \textcolor{colorMonroe}{\textbf{0.730}} & 0.729 & \textcolor{colorMiniMolPFN}{\textbf{0.716}} & \textcolor{colorMiniMol}{\textbf{0.672}} & 0.654 & 0.651 & \textcolor{colorCheMeleonPFN}{\textbf{0.651}} & 0.632 & \textcolor{colorCheMeleon}{\textbf{0.609}} & \textcolor{colorMolFormer}{\textbf{0.580}} &  &  \\
Pgp Broccatelli & 973 & AUROC ($\uparrow$) & 0.954 & \textcolor{colorMiniMolPFN}{\textbf{0.946}} & \textcolor{black}{\textbf{0.944}} & \textcolor{colorMonroe}{\textbf{0.942}} & 0.942 & \textcolor{colorMiniMol}{\textbf{0.941}} & 0.938 & 0.935 & 0.930 & 0.929 & 0.929 & 0.928 & 0.926 & 0.923 & 0.922 \\
Clearance Hepatocyte AZ & 970 & Spearman ($\uparrow$) & 0.538 & 0.536 & 0.525 & \textcolor{black}{\textbf{0.503}} & 0.498 & 0.466 & \textcolor{colorMiniMolPFN}{\textbf{0.463}} & \textcolor{colorMonroe}{\textbf{0.452}} & 0.447 & 0.440 & 0.440 & 0.439 & 0.431 & \textcolor{colorCheMeleonPFN}{\textbf{0.431}} & 0.430 \\
VDss Lombardo & 904 & Spearman ($\uparrow$) & 0.713 & 0.707 & \textcolor{black}{\textbf{0.700}} & \textcolor{colorMonroe}{\textbf{0.700}} & \textcolor{colorCheMeleon}{\textbf{0.684}} & \textcolor{colorMiniMolPFN}{\textbf{0.680}} & \textcolor{colorCheMeleonPFN}{\textbf{0.639}} & 0.637 & 0.628 & 0.627 & \textcolor{colorMolFormer}{\textbf{0.615}} & 0.609 & 0.600 & 0.582 & 0.561 \\
Clearance Microsome AZ & 881 & Spearman ($\uparrow$) & 0.683 & \textcolor{black}{\textbf{0.680}} & \textcolor{colorMonroe}{\textbf{0.678}} & 0.653 & 0.630 & \textcolor{colorMiniMolPFN}{\textbf{0.628}} & 0.626 & 0.625 & 0.625 & \textcolor{colorCheMeleon}{\textbf{0.620}} & \textcolor{colorCheMeleonPFN}{\textbf{0.607}} & 0.599 & 0.597 & 0.586 & 0.585 \\
Caco2 Wang & 728 & MAE ($\downarrow$) & 0.256 & \textcolor{black}{\textbf{0.257}} & \textcolor{colorMonroe}{\textbf{0.274}} & 0.274 & 0.276 & 0.282 & 0.285 & 0.287 & 0.287 & 0.289 & 0.297 & \textcolor{colorCheMeleon}{\textbf{0.302}} & 0.316 & 0.319 & 0.321 \\
HIA Hou & 578 & AUROC ($\uparrow$) & \textcolor{colorMiniMolPFN}{\textbf{0.996}} & \textcolor{black}{\textbf{0.994}} & 0.990 & 0.989 & 0.988 & \textcolor{colorMiniMol}{\textbf{0.987}} & 0.986 & \textcolor{colorCheMeleon}{\textbf{0.985}} & 0.985 & 0.981 & 0.981 & 0.978 & \textcolor{colorMonroe}{\textbf{0.978}} & 0.975 & 0.974 \\
CYP3A4 Substrate & 535 & AUROC ($\uparrow$) & 0.696 & \textcolor{colorMonroe}{\textbf{0.691}} & 0.685 & \textcolor{black}{\textbf{0.685}} & 0.667 & \textcolor{colorMiniMol}{\textbf{0.664}} & 0.662 & 0.655 & \textcolor{colorMiniMolPFN}{\textbf{0.655}} & 0.650 & 0.647 & 0.640 & \textcolor{colorCheMeleonPFN}{\textbf{0.640}} & 0.639 & 0.633 \\
CYP2C9 Substrate & 534 & AUPRC ($\uparrow$) & 0.478 & \textcolor{colorMonroe}{\textbf{0.462}} & \textcolor{black}{\textbf{0.457}} & \textcolor{colorMiniMol}{\textbf{0.444}} & 0.441 & 0.437 & 0.437 & 0.433 & \textcolor{colorMiniMolPFN}{\textbf{0.432}} & \textcolor{colorCheMeleonPFN}{\textbf{0.428}} & 0.417 & 0.417 & 0.415 & 0.407 & 0.400 \\
CYP2D6 Substrate & 532 & AUPRC ($\uparrow$) & 0.736 & \textcolor{colorMonroe}{\textbf{0.735}} & 0.731 & \textcolor{black}{\textbf{0.727}} & \textcolor{colorMiniMolPFN}{\textbf{0.726}} & 0.720 & 0.713 & 0.711 & 0.704 & 0.704 & 0.686 & 0.685 & 0.680 & 0.677 & \textcolor{colorMiniMol}{\textbf{0.673}} \\
Half Life Obach & 532 & Spearman ($\uparrow$) & \textcolor{colorMiniMolPFN}{\textbf{0.639}} & 0.625 & \textcolor{colorMonroe}{\textbf{0.617}} & \textcolor{black}{\textbf{0.604}} & 0.576 & 0.573 & 0.562 & 0.557 & 0.547 & 0.544 & \textcolor{colorCheMeleonPFN}{\textbf{0.532}} & 0.485 & \textcolor{colorCheMeleon}{\textbf{0.469}} & 0.438 & 0.392 \\
hERG & 523 & AUROC ($\uparrow$) & 0.891 & 0.880 & 0.875 & 0.874 & 0.871 & \textcolor{black}{\textbf{0.864}} & 0.856 & \textcolor{colorMonroe}{\textbf{0.856}} & 0.855 & \textcolor{colorMiniMolPFN}{\textbf{0.845}} & 0.841 & 0.840 & 0.836 & 0.825 & \textcolor{colorCheMeleonPFN}{\textbf{0.823}} \\
Bioavailability Ma & 512 & AUROC ($\uparrow$) & 0.748 & 0.742 & 0.730 & 0.729 & 0.728 & \textcolor{black}{\textbf{0.715}} & 0.714 & \textcolor{colorMiniMolPFN}{\textbf{0.714}} & 0.706 & \textcolor{colorMonroe}{\textbf{0.699}} & \textcolor{colorMolFormer}{\textbf{0.684}} & 0.675 & 0.673 & 0.672 & 0.667 \\
DILI & 379 & AUROC ($\uparrow$) & \textcolor{black}{\textbf{0.952}} & \textcolor{colorMiniMolPFN}{\textbf{0.951}} & 0.933 & \textcolor{colorMiniMol}{\textbf{0.931}} & 0.925 & 0.925 & \textcolor{colorMonroe}{\textbf{0.924}} & 0.919 & 0.919 & 0.917 & 0.909 & 0.906 & \textcolor{colorCheMeleonPFN}{\textbf{0.901}} & 0.899 & 0.887 \\
Fang HPPB & 126 & Pearson ($\uparrow$) & 0.888 & 0.884 & \textcolor{black}{\textbf{0.853}} & \textcolor{colorMonroe}{\textbf{0.822}} & 0.815 & \textcolor{colorMiniMolPFN}{\textbf{0.787}} & 0.774 & \textcolor{colorCheMeleon}{\textbf{0.774}} & \textcolor{colorCheMeleonPFN}{\textbf{0.765}} & \textcolor{colorMolFormer}{\textbf{0.736}} & \textcolor{colorMiniMol}{\textbf{0.656}} & 0.636 & 0.635 &  &  \\
Fang RPPB & 111 & Pearson ($\uparrow$) & 0.908 & 0.886 & \textcolor{colorMonroe}{\textbf{0.860}} & \textcolor{black}{\textbf{0.847}} & \textcolor{colorCheMeleonPFN}{\textbf{0.776}} & \textcolor{colorMiniMolPFN}{\textbf{0.774}} & 0.747 & \textcolor{colorMolFormer}{\textbf{0.730}} & \textcolor{colorCheMeleon}{\textbf{0.709}} & 0.680 & 0.614 & \textcolor{colorMiniMol}{\textbf{0.487}} &  &  &  \\
\bottomrule
\end{tabular}
}
}
\end{table*}

\clearpage
\subsection{Per-task pairwise comparisons}
\label{app:per-task-grids}

The win-rate counts reported in the main text aggregate the same pairwise tests applied per task. For visual inspection of those underlying tests, we provide one panel per task showing each method's mean and 95\% confidence interval (5 seeds per method). Confidence intervals use the pooled within-group variance from the equal-variance assumption, so within a panel all error bars share the same length. The dashed vertical lines mark the 95\% CI of the best method (the practical-equivalence band). The set of methods shown in \textcolor[HTML]{1F4FFF}{blue}, \textcolor[HTML]{888888}{gray}, or \textcolor[HTML]{E8A33D}{amber} on a given task is exactly the set that contributes to that task's win count.

\begin{figure}[!htbp]
\centering
\includegraphics[width=\textwidth]{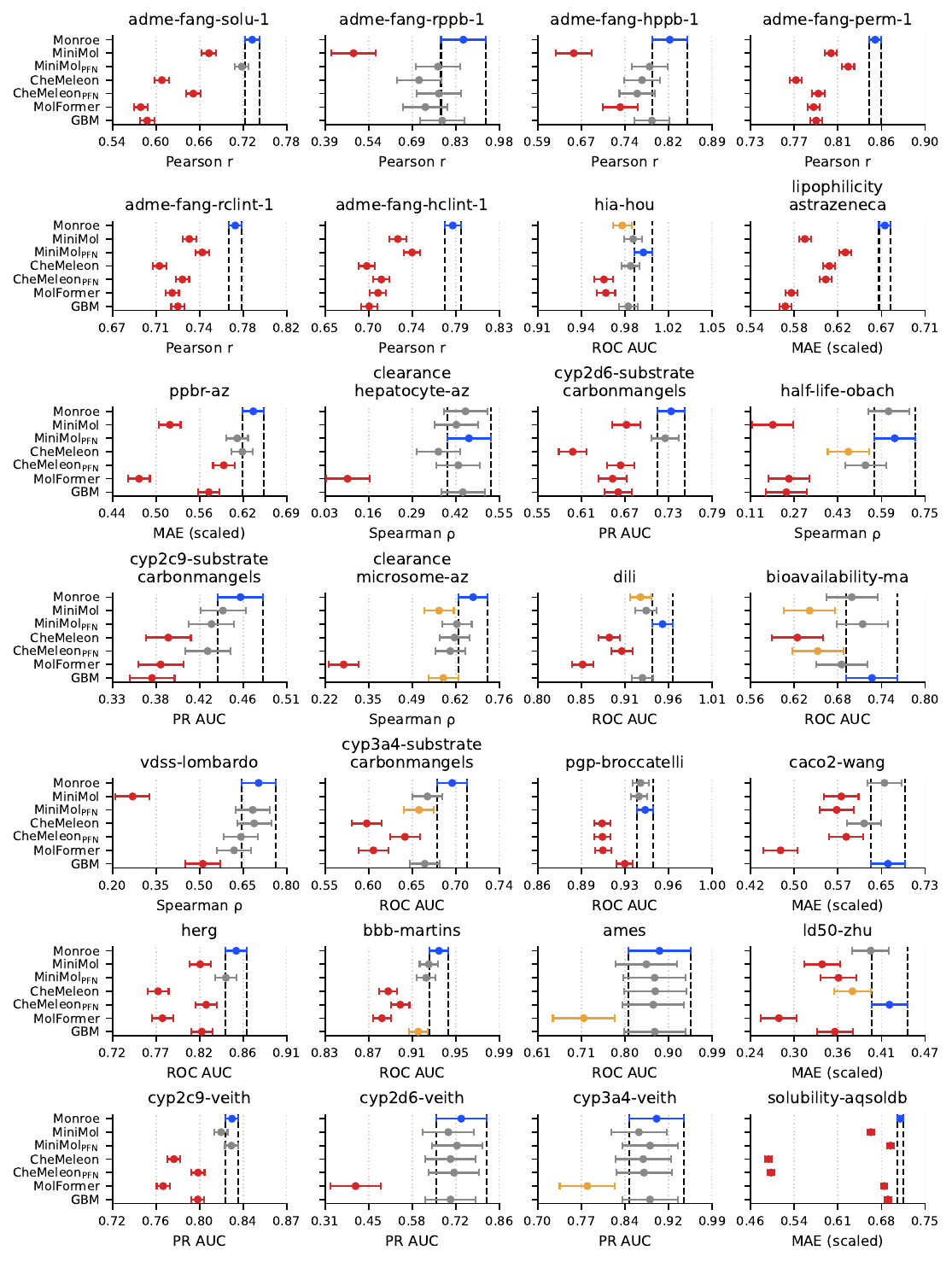}
\caption{Per-task pairwise comparisons on 28 ADMET tasks from Polaris.
Error bars are 95\% confidence intervals; interval overlap corresponds to an \emph{uncorrected} pairwise comparison. The colour reflects the multiplicity-corrected verdict against the best method (Tukey HSD followed by Benjamini--Hochberg FDR): \textcolor[HTML]{1F4FFF}{\textbf{blue}} marks the best-mean method on the task (a winner); \textcolor[HTML]{888888}{\textbf{gray}} marks methods tied with the best (also winners, counted as ties for first) whose interval overlaps the best's; \textcolor[HTML]{E8A33D}{\textbf{amber}} marks methods that are also counted as winners under FDR correction but whose interval does \emph{not} overlap the best's, meaning the intervals separate but the correction still classifies them as tied; \textcolor[HTML]{D62728}{\textbf{red}} marks methods significantly worse than the best (losers).
Dashed vertical lines mark the best method's 95\% CI; all error bars in a panel are equal-length under the pooled-variance assumption.
MAE-based tasks are reported as scaled MAE ($1 - \mathrm{MAE}/\mathrm{scale}$, see \cite{walters2024polaris}) so that higher is better across all panels.}
\label{fig:per-task-polaris}
\end{figure}

\begin{figure}[!htbp]
\centering
\includegraphics[width=\textwidth]{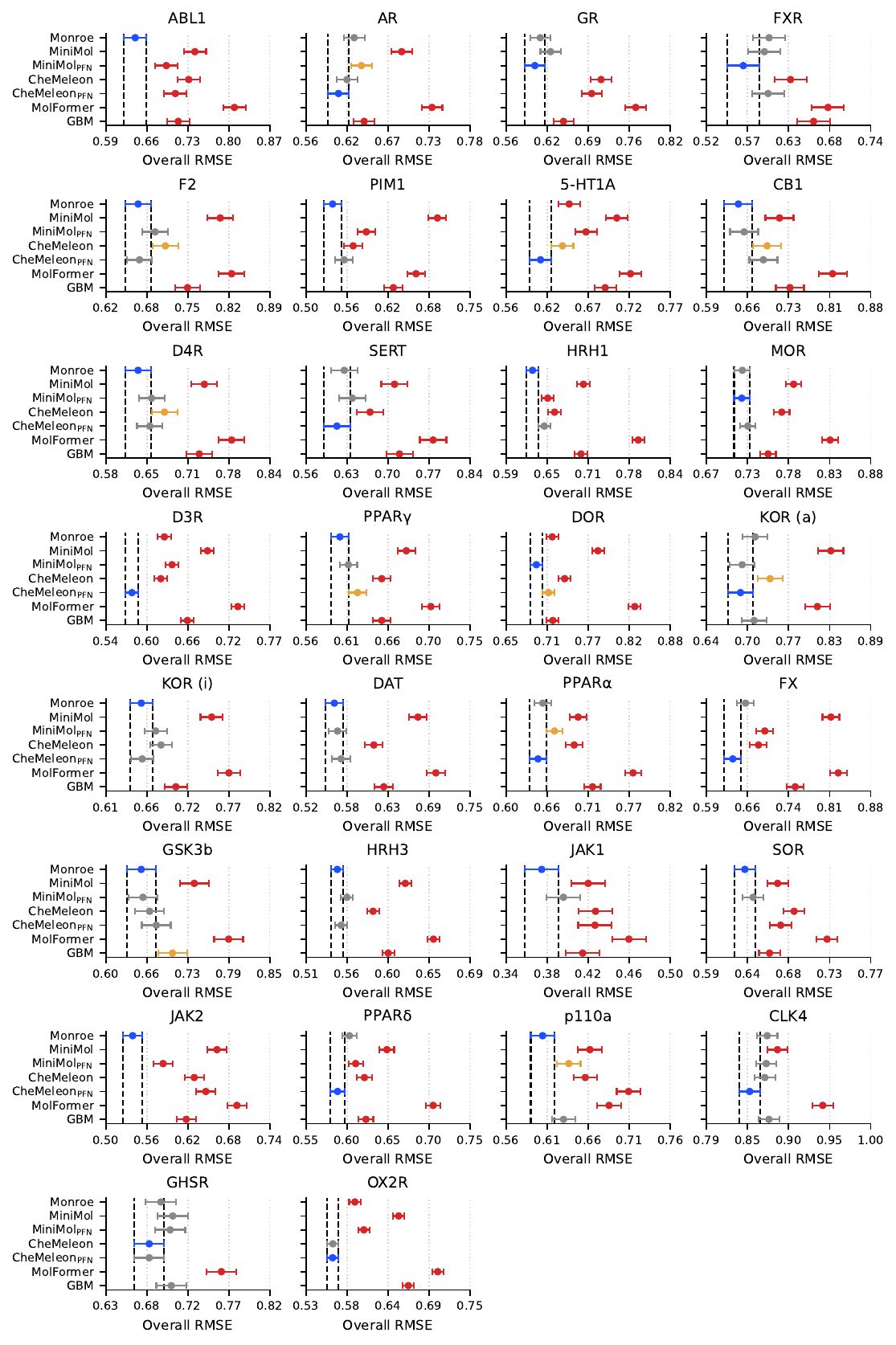}
\caption{Per-task pairwise comparisons on 30 activity-cliff tasks from MoleculeACE.
Same color and CI conventions as \cref{fig:per-task-polaris}: \textcolor[HTML]{1F4FFF}{\textbf{blue}} = best on this task, \textcolor[HTML]{888888}{\textbf{gray}} = tied with the best and interval overlaps, \textcolor[HTML]{E8A33D}{\textbf{amber}} = tied with the best under FDR correction but interval does not overlap, \textcolor[HTML]{D62728}{\textbf{red}} = significantly worse than the best.
Each panel reports the overall test RMSE on one dataset (lower is better, so the best-mean method has the smallest RMSE).}
\label{fig:per-task-moleculeace}
\end{figure}

\clearpage
\section{Implementation details}
\label{sec:implementation-details}
The architecture of the encoder is summarized below:
{\tiny
\begin{lstlisting}[language=Python]
========================================================================================================================
ENCODER ARCHITECTURE
========================================================================================================================
  GritTransformer(
  (edge_rbf): RBF()
  (node_oh): ModuleDict(
    (valence_implicit): Embedding(14, 128)
    (valence_total): Embedding(14, 128)
    (hybridization): Embedding(9, 128)
    (atomic_number): Embedding(44, 128)
    (total_num_hs): Embedding(10, 128)
    (is_aromatic): Embedding(2, 128)
    (is_in_ring): Embedding(2, 128)
    (chirality): Embedding(3, 128)
    (period): Embedding(8, 128)
    (degree): Embedding(14, 128)
    (group): Embedding(20, 128)
  )
  (edge_oh): ModuleDict(
    (conjugation): Embedding(2, 128)
    (is_in_ring): Embedding(2, 128)
    (bond_type): Embedding(5, 128)
    (stereo): Embedding(9, 128)
  )
  (node_proj): Linear(in_features=156, out_features=720, bias=True)
  (edge_proj): Linear(in_features=180, out_features=720, bias=True)
  (rwpe_node_bias_proj): Linear(in_features=20, out_features=720, bias=False)
  (rwpe_edge_bias_proj): Linear(in_features=20, out_features=720, bias=False)
  (grit_layers): ModuleList(
    (0-9): 10 x GritTransformerLayer(in_channels=720, out_channels=720, heads=10)
    [GritTransformerLayer(
      (attention): MultiHeadAttentionLayerGritSparse(
        (dropout): Dropout(p=0.0, inplace=False)
        (Q): Linear(in_features=720, out_features=720, bias=True)
        (K): Linear(in_features=720, out_features=720, bias=True)
        (E): Linear(in_features=720, out_features=1440, bias=True)
        (V): Linear(in_features=720, out_features=720, bias=True)
      )
      (O_h): Linear(in_features=720, out_features=720, bias=True)
      (O_e): Linear(in_features=720, out_features=720, bias=True)
      (batch_norm1_h): BatchNorm1d(720, eps=1e-05, momentum=0.1, affine=True, track_running_stats=True)
      (batch_norm1_e): BatchNorm1d(720, eps=1e-05, momentum=0.1, affine=True, track_running_stats=True)
      (batch_norm2_h): BatchNorm1d(720, eps=1e-05, momentum=0.1, affine=True, track_running_stats=True)
      (FFN_h_layer1): Linear(in_features=720, out_features=1440, bias=True)
      (FFN_h_layer2): Linear(in_features=1440, out_features=720, bias=True)
    )]
  )
  (pooling): MultiHeadAttentionPooling(
    (gate_dropout): Dropout(p=0.05, inplace=False)
    (gate): Linear(in_features=720, out_features=10, bias=True)
    (value_proj): Linear(in_features=720, out_features=720, bias=False)
  )
)
========================================================================================================================
PARAMETER COUNT
========================================================================================================================
Node embeddings:           17.9K
Edge embeddings:            2.3K
Input projections:        272.2K
Transformer layers:       57.68M
Pooling:                  525.6K
------------------------------
TOTAL:                    58.50M
\end{lstlisting}
}

\label{app:code}
\methodname is implemented in PyTorch with minimal dependencies,
and adapts methods from LibMTL~\citep{lin2023libmtl}.
Optimization of the MTL objective using Adam on 81 million molecules took 26 hours on four H100 GPUs.
The entire preprocessing, training, and evaluation pipelines are publicly available.  A focus on efficiency of implementation means that it is practical for other researchers to consider adaptations of the pretraining losses and architectures in future research.
The code is available at: \href{https://github.com/blazejba/monroe}{github.com/blazejba/monroe}.

\subsection{Architectural details}
\label{sec:architectural-details}
\label{sec:featurization-details}

\paragraph{GRIT layer update.}
For each layer and head $h$, given nodes $\vxi \in \mathbb R^{d_h}$ and edges $\eij \in \mathbb R^{d_e}$, our sparse GRIT layer produces $\vxi^{+}$ and $\eij^{+}$ via
\begin{equation}
\def\qi{\vv q_i^h}
\def\kj{\vv k_j^h}
\def\vj{\vv v_j^h}
\begin{aligned}
\qi, ~\kj, ~\vj &= W_Q^h \vxi, ~~W_K^h \vxj, ~~W_V^h \vxj, \\
\sijh &= \varphi\!\big((\qi \odot \kj)\odot W_E^h \eij\big) + W_{Eb}^h \eij, \\
\aijh &= \mathrm{softmax}_{j\in\mathcal{N}(i)}\!\left(d_h^{-1/2}\cdot{(\vv w_{\text{attn}}^h)^\top \sijh}\right), \\
\mih &= \sum_{j\in\mathcal{N}(i)} \aijh \big( \vj + W_{ev}^h \sijh \big), \\
\vxi^{+}, ~\eij^{+} &= W_O \big[ \|_{h=1}^H \mih \big], ~~ W_{Eo} \big[ \|_{h=1}^H \sijh \big],
\end{aligned}
\end{equation}
with $\varphi(z)=\mathrm{sign}(z)\sqrt{|z|+\epsilon}$ and learned attention vectors $\vv w_{\text{attn}}^h \in \mathbb{R}^{d_h}$.
We then apply degree-based scaling $\vv x_i^{+}\leftarrow \lambda_1 \vv x_i^{+} + \lambda_2 \vv x_i^{+}\log(\deg_i+1)$, residual connections, batch normalization, and a two-layer FFN. The neighborhood $\mathcal{N}(i)$ is restricted to observed (and stereochemistry-augmented) edges plus the virtual-node edge, giving $O(|E|)$ attention rather than the dense $O(|V|^2)$ used in the original GRIT.

\paragraph{Node features.} Each atom $v$ has two groups of features: (i) eight continuous scalars:
number of radical electrons, electronegativity, first ionization energy,
covalent radius, melting point, formal charge, van der Waals radius, and atomic
weight; and (ii) eleven categorical codes:
implicit valence, total valence, hybridization, element symbol, total number of
attached hydrogens, aromaticity, ring membership, chirality (R/S),
period, degree, and group. Each categorical feature has a fixed vocabulary, with
unseen values mapped to a dedicated \emph{other} category.

\paragraph{Edge features.} Each bond $(u,v)$ is assigned four categorical codes: conjugation,
ring membership, bond type (single/double/triple/aromatic), and stereochemistry
(none/any/E/Z/cis/trans). The base molecular graph is treated as undirected (each bond is inserted in both directions). When stereochemical augmentation is enabled, additional edges are inserted to encode E/Z double bonds and R/S chiral centers; these edges use two auxiliary stereo categories and map the remaining bond attributes to the \emph{other} category. Unlike base bonds, the R/S radial and cyclic edges (Eq.~5) are inserted in a single direction so that the cycle orientation $a{\to}b{\to}c{\to}a$ vs.\ $a{\to}c{\to}b{\to}a$ is preserved as topology, allowing the encoder to distinguish enantiomers.

\paragraph{Virtual node and degree scaling.} Beyond the node and edge inputs constructed in Eqs.~(1)--(3), each graph receives one extra virtual node with a dedicated categorical token and zero continuous/positional features, plus edges connecting every atom to this virtual node. Within each layer, $\log(\deg+1)$ is additionally used as a degree-based scaling signal applied to the GRIT layer output.

\subsection{Stereochemistry details}
\label{sec:stereo-details}

Stereochemistry refers to the three-dimensional arrangement of atoms in molecules.
Two molecules with identical connectivity but different spatial arrangements are
\emph{stereoisomers}. This section explains the two key types of stereoisomerism
that our model encodes.

\paragraph{CIP priorities.}
The Cahn--Ingold--Prelog (CIP) rules provide a consistent ordering of substituents:
(1)~rank by atomic number of the directly attached atom (higher = higher priority);
(2)~for ties, compare next atoms outward until a difference is found;
(3)~treat double bonds as duplicated single bonds (e.g., C=O counts as C bonded to two oxygens).
This ordering underlies both R/S and E/Z nomenclature.

\paragraph{R/S chirality (tetrahedral stereocenters).}
A carbon with four \emph{distinct} substituents is a stereocenter. After ranking
substituents as $a{>}b{>}c{>}d$ by CIP priority, orient the molecule so that
the lowest-priority group $d$ points away from you (like looking down an axle
toward the back wheel of a car).
Trace $a{\to}b{\to}c$: clockwise defines \textbf{R} (Latin \emph{rectus}, right),
counterclockwise defines \textbf{S} (Latin \emph{sinister}, left).
Enantiomers---non-superimposable mirror images---have opposite R/S configurations
at every stereocenter.

\paragraph{E/Z configuration (geometric isomers).}
Double bonds prevent free rotation, so substituents on each carbon can be locked
on the same or opposite sides. For each double-bond carbon, identify the
higher-priority substituent via CIP.
\textbf{Z} (German \emph{zusammen}, together): both higher-priority substituents
lie on the \emph{same} side.
\textbf{E} (German \emph{entgegen}, opposite): higher-priority substituents lie on
\emph{opposite} sides.
E and Z isomers are diastereomers---stereoisomers that are \emph{not} mirror images.

\paragraph{Why invariant encoders need augmentation.}
A rotation-invariant encoder maps 3D coordinates to scalar distances, erasing
chirality: a molecule and its mirror image produce identical distance matrices.
Similarly, E/Z bond geometry may be noisy or absent in approximate conformers.
Our auxiliary edges convert stereochemical labels into \emph{graph topology}:
\begin{itemize}
  \item \textbf{E/Z:} ``parallel'' edges link same-side substituents (stereo-defining pair);
        ``diagonal'' edges link opposite-side substituents (contrastive pair).
        Labeling one set with the observed configuration and the other with its
        complement ensures the network receives a unique signal for each isomer.
  \item \textbf{R/S:} radial edges from the lowest-priority substituent $d$ to the
        others, plus a directed cycle $a{\to}b{\to}c{\to}a$ that encodes the S
        configuration (counterclockwise when viewed from $d$); the reversed cycle
        $a{\to}c{\to}b{\to}a$ encodes R.
\end{itemize}
Because these signals are discrete edge labels rather than continuous distances,
they survive rotation invariance and let message passing distinguish stereoisomers
that would otherwise be indistinguishable.

\subsection{PM6 dataset details}
\label{sec:pm6-details}
PM6 (Parametric Method 6) is a semi-empirical quantum chemistry method
that solves an approximation of the Schrödinger equation \citep{stewart2007optimization}.
Unlike \emph{ab initio} Density Functional Theory (DFT) methods such as B3LYP,
which calculate electron interactions explicitly, PM6 uses empirical parameters derived from experimental data to simplify the Hamiltonian.
This approximation makes PM6 orders of magnitude faster than DFT,
enabling calculations on the scale of tens of millions of molecules---a scale that would be computationally prohibitive with higher-level theory.
Despite its lower precision, \citet{Nakata2020PubChemQC} showed that PM6-optimized geometries
and electronic properties (e.g., HOMO/LUMO energies) exhibit strong correlation ($R^2 \approx 0.9$)
with those obtained from B3LYP/6-31G* calculations, providing a high-quality signal for large-scale pre-training.

We use the PM6\_83M dataset derived from the PubChemQC project \citep{Nakata2020PubChemQC}, as curated by \citet{Beaini2024Towards} for the UltraLarge benchmark.
The raw release contains 83 million unique molecules; after deduplication, removal of molecules whose InChIKey overlaps the test split of any downstream benchmark, and discarding featurization failures, we retain roughly 81 million molecules for pretraining---the figure cited throughout the paper.
Since calculations are performed for four different molecular states---ground state (S0), lowest energy triplet excited state (T0), cation, and anion---the raw dataset comprises a total of 221M semi-empirical PM6 computations.
This represents the largest available dataset for molecular representation learning in terms of unique chemical structures.

The pre-training objective uses 62 graph-level targets per molecule. These include quantum properties like Alpha HOMO, Alpha HOMO-LUMO gap, Beta HOMO, and Total energy across the available states (S0, T0, cation, anion).
Additionally, we predict 3D descriptors computed from the PM6-optimized conformations, including Spherocity, Plane of best fit, and Principal length along the three main axes.
Node-level targets (atomic Mulliken charges and electronic spins across the four states) are also available in the source data, but we observed that including them degraded downstream performance under our training conditions, so they are excluded from training.

\subsubsection{Loss functions for heterogeneous PM6 targets}
\label{subsec:pm6-losses}
The PM6 objective spans heterogeneous quantities with different supports and error structures: some targets are signed and live on $\mathbb{R}$ (e.g., energies, orbital levels, Mulliken charges), others are strictly positive scale parameters (e.g., lengths and molecular size descriptors), others are bounded ratios in $[0,1]$ (e.g., spherocity, QED, FSP3), and several are small bounded integer counts (e.g., ring counts). A single regression loss is therefore a poor inductive bias; instead, we choose the loss \emph{per target} to match (i) the target's natural domain constraints and (ii) the most appropriate notion of error (absolute vs.\ relative, additive vs.\ multiplicative, continuous vs.\ discrete).

\vspace{0.5em}
\paragraph{Huber loss for signed real-valued targets.}
For signed targets where an \emph{absolute} deviation is meaningful, we use the Huber loss. Let $r = \hat y - y$ denote the residual. With threshold $\delta>0$,
\begin{equation}
\mathcal{L}_{\text{Huber}}(y,\hat y;\delta)=
\begin{cases}
\frac{1}{2}r^2, & |r|\le \delta,\\
\delta\left(|r|-\frac{1}{2}\delta\right), & |r|>\delta.
\end{cases}
\end{equation}
PM6-derived targets can contain rare heavy-tailed errors. Huber preserves the benefits of MSE near the optimum while limiting the influence of outliers, which is particularly important in a multi-target setting.

\vspace{0.5em}
\paragraph{Log-Huber loss for strictly positive scale-like targets.}
For strictly positive quantities where \emph{relative} error is the natural scale (e.g., lengths and size descriptors), we apply Huber in log-space. Using a small $\varepsilon>0$, the network output $\hat y$ is interpreted as a log-domain prediction and compared to the log-transformed target $y^{(\log)} = \log(y+\varepsilon)$:
\begin{equation}
\mathcal{L}_{\text{log-Huber}}(y,\hat y;\delta,\varepsilon)
= \mathcal{L}_{\text{Huber}}\!\left(y^{(\log)}, \hat y;\delta\right).
\end{equation}
Log-space converts multiplicative/scale-dependent noise into approximately additive noise, making optimization more uniform across magnitudes and ensuring positivity after exponentiation.

\vspace{0.5em}
\paragraph{Log-ratio squared loss for spherocity.}
Spherocity is a bounded shape descriptor ($[0,1]$) defined through ratios of geometric quantities. We therefore penalize squared error in log-ratio space:
\begin{equation}
\mathcal{L}_{\text{log-ratio}}(s,\hat s;\varepsilon)
= \left(\log(\hat s+\varepsilon) - \log(s+\varepsilon)\right)^2,
\qquad s \in [0,1].
\end{equation}
The log transform makes multiplicative deviations symmetric (e.g., $\times 2$ and $\times \frac12$), which aligns with the ratio nature of geometric descriptors and yields stable gradients.

\vspace{0.5em}
\paragraph{Beta negative log-likelihood for bounded ratios/scores.}
For bounded ratios/scores in $(0,1)$, we use a Beta likelihood and clip targets to avoid boundary issues:
\begin{equation}
\tilde y = \mathrm{clip}(y,\varepsilon,1-\varepsilon).
\end{equation}
We parameterize $\mathrm{Beta}(\alpha,\beta)$ via mean $\mu \in (0,1)$ and concentration $\kappa>0$:
\begin{equation}
\begin{aligned}
\mu &= \mathrm{clip}(\sigma(z_\mu),0.01,0.99),\qquad
\kappa = \mathrm{clip}(\mathrm{softplus}(z_\kappa)+2,2,200),\\
\alpha &= \mu\kappa,\qquad \beta = (1-\mu)\kappa,
\end{aligned}
\end{equation}
and compute the negative log-likelihood
\begin{equation}
\mathcal{L}_{\text{BetaNLL}}(\tilde y;\alpha,\beta)
= -(\alpha-1)\log \tilde y - (\beta-1)\log(1-\tilde y)
+ \log \mathrm{B}(\alpha,\beta),
\end{equation}
where $\log \mathrm{B}(\alpha,\beta)=\log\Gamma(\alpha)+\log\Gamma(\beta)-\log\Gamma(\alpha+\beta)$.
Since Beta models a \emph{density} on $(0,1)$, $\mathcal{L}_{\text{BetaNLL}}$ can be negative when the predicted distribution is highly concentrated near $\tilde y$.
For numerical stability and to ensure a bounded loss range, we clamp the concentration parameters,
\begin{equation}
\alpha \leftarrow \mathrm{clip}(\alpha,\alpha_{\min},\alpha_{\max}),\qquad
\beta \leftarrow \mathrm{clip}(\beta,\beta_{\min},\beta_{\max}),
\end{equation}
with $\alpha_{\min}=\beta_{\min}=1$, $\alpha_{\max}=\beta_{\max}=100$, and target-clipping threshold $\varepsilon=10^{-4}$. Clamping bounds the most negative achievable NLL under our parameterization. To obtain a positive task loss for uncertainty weighting, we apply softplus to the mean Beta NLL over the observed labels in the batch:
\begin{equation}
\mathcal{L}_{\mathrm{Beta}} = \mathrm{softplus}\!\left(\overline{\mathcal{L}}_{\mathrm{BetaNLL}}\right).
\end{equation}

\vspace{0.5em}
\paragraph{Ordinal loss for bounded integer counts.}
For bounded counts, we use an ordinal formulation that preserves ordering. Let $y\in\{m,m+1,\dots,M\}$ and define $y' = y-m \in \{0,1,\dots,K\}$ with $K=M-m$. For $k=1,\dots,K$, define $t_k=\mathbbm{1}[y' \ge k]$. The model outputs logits $a_k$ and predicts $p_k=\sigma(a_k)\approx \Pr(y' \ge k)$. The ordinal loss sums binary cross-entropies:
\begin{equation}
\mathcal{L}_{\text{ord}}(y,\{a_k\}_{k=1}^K)
= \sum_{k=1}^K \mathrm{BCE}\!\left(t_k,\sigma(a_k)\right).
\end{equation}
These targets are discrete and bounded, and the ordinal construction respects their ordering, penalizing near-miss errors less than large deviations while avoiding the mismatch of unconstrained real-valued regression.

\subsection{PCBA dataset details}
\label{sec:pcba-details}
PCBA \citep{Beaini2024Towards} is the standard PubChem BioAssay benchmark, comprising 1{,}328 binary high-throughput-screen assays measured on 1.56\,M small molecules. Each (molecule, assay) cell is either active (1), inactive (0), or missing (NaN); the matrix is sparse---most molecules have been screened against only a handful of the 1{,}328 assays.

\paragraph{Preprocessing and benchmark contamination.}
We canonicalize each PCBA SMILES to InChI, deduplicate, and route any molecule whose InChI also appears in a Polaris or MoleculeACE \emph{test} split into our validation set so it cannot leak into pretraining. The remaining molecules form the PCBA training pool.

\begin{figure}[!htbp]
\centering
\includegraphics[width=\textwidth]{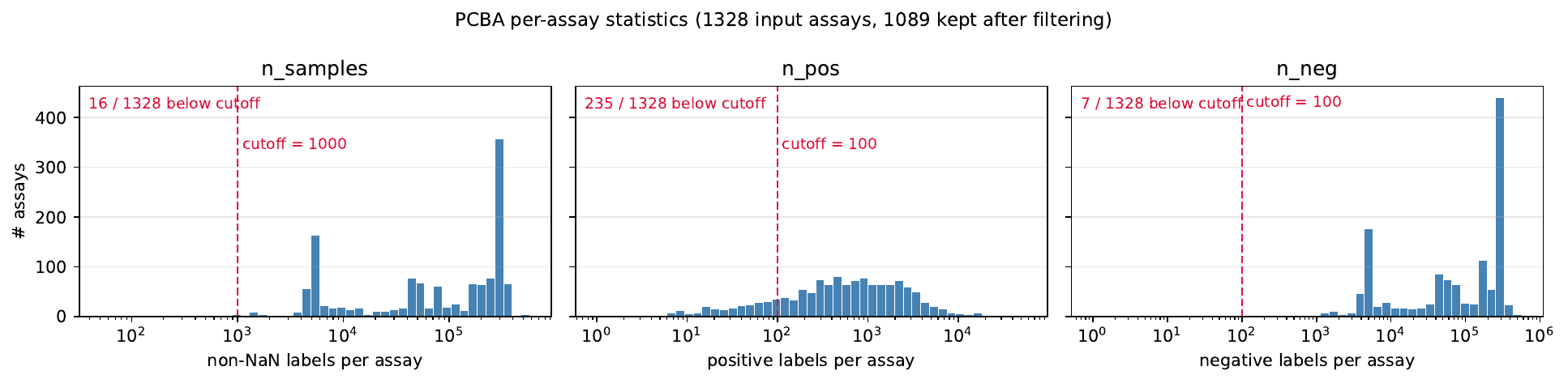}
\caption{PCBA per-assay distributions of the three filter signals: number of non-NaN labels (left), positive labels (middle), and negative labels (right) per assay. Dashed crimson lines mark the training-time cutoffs (1000 / 100 / 100). An assay is dropped if it fails any single threshold; this leaves 1089 of the 1328 input assays. The positive-count threshold accounts for the bulk of the rejected assays (235 below the cutoff), reflecting strong class imbalance in high-throughput screens.}
\label{fig:pcba_assay_distributions}
\end{figure}

\paragraph{Per-assay filtering.}
Computed over the training rows only, we drop any assay that does not satisfy \emph{all three} of the following: at least 100 positive labels, at least 100 negative labels, and at least 1{,}000 total non-NaN labels. \cref{fig:pcba_assay_distributions} shows the three distributions and the cutoffs as dashed lines. The thresholds remove assays that are either too small or too class-skewed to provide a useful learning signal, leaving 1089 of the 1328 input assays. The dominant cause of rejection is the positive-count threshold (235 assays below the cutoff of 100 positives), reflecting the strong class imbalance of high-throughput-screen data.

\paragraph{Loss and head architecture.}
Even after filtering, surviving assays have an average positive rate of roughly $1\!:\!100$. A naive binary cross-entropy is dominated by trivially-classified negatives, so we use focal loss \citep{lin2017focal} with $\gamma=2$, $\alpha=0.25$, which downweights well-classified examples and concentrates gradient on hard ones.
The PCBA head is a single linear probe ($\mathrm{Linear}(d, K)$ with $K=1089$, no hidden layers) that emits one logit per surviving assay from the pooled fingerprint. The forward pass slices this $[B, K]$ output into $K$ per-assay views; each view is registered as its own task in the MTL bookkeeping with its own focal-BCE loss (averaged over non-NaN cells of that assay) and its own learnable $\sigma_t$ in UW. PCBA therefore contributes $K=1089$ independent terms to the UW sum, yielding a total of $T=1152$ tasks (62 PM6 + 1 conformer denoising + 1089 PCBA). The head weights themselves are shared, but the task-level weighting is per-assay, which lets UW automatically downweight assays whose loss is dominated by noise.

\clearpage
\section{MoleculeACE details}
\label{app:moleculeace}

\begin{figure*}[!ht]
\centering
\begin{minipage}[t]{0.48\textwidth}
\centering
\vspace{0pt}
\captionof{figure}{The activity distributions for cliff and non-cliff molecules in MoleculeACE.}
\includegraphics[width=\linewidth,trim={0.2cm 0.2cm 0.2cm 0.2cm},clip=true]{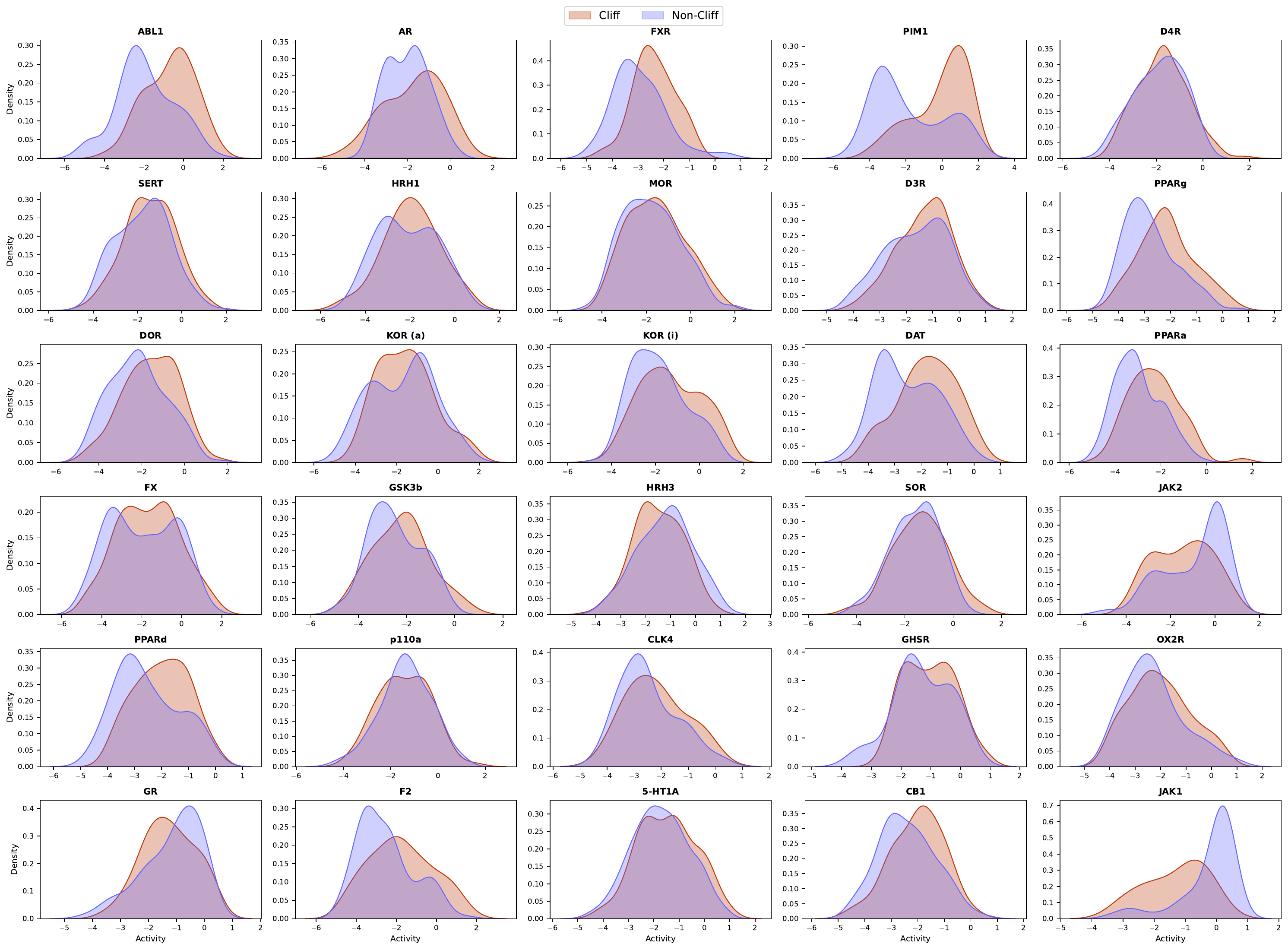}
\label{fig:moleculeace_activity_dist}
\end{minipage}
\hfill
\begin{minipage}[t]{0.48\textwidth}
\centering
\vspace{0pt}
\captionof{table}{Overview of MoleculeACE dataset statistics, highlighting the prevalence of stereochemical features in the test set.}
\label{tab:moleculeace_stats}
\resizebox{\linewidth}{!}{
\begin{tabular}{llcccc}
\toprule
\textbf{Target Name} & \textbf{ID} & \textbf{Cliff Mols} & \textbf{Non-Cliff Mols} & \textbf{R/S Stereo (\%)} & \textbf{E/Z Stereo (\%)} \\
\midrule
Serotonin 1a receptor & 5-HT1A & 245 & 421 & 35.1 & 2.3 \\
Tyrosine-protein kinase ABL1 & ABL1 & 67 & 94 & 21.1 & 1.2 \\
Androgen Receptor & AR & 32 & 102 & 67.9 & 6.7 \\
Cannabinoid receptor 1 & CB1 & 76 & 132 & 33.7 & 10.1 \\
Dual specificity protein kinase CLK4 & CLK4 & 13 & 136 & 14.8 & 0.0 \\
Dopamine D3 receptor & D3R & 320 & 413 & 27.3 & 4.2 \\
Dopamine D4 receptor & D4R & 148 & 226 & 29.4 & 5.1 \\
Dopamine transporter & DAT & 55 & 158 & 72.8 & 0.9 \\
Delta opioid receptor & DOR & 204 & 317 & 82.1 & 5.8 \\
Thrombin & F2 & 220 & 333 & 54.8 & 3.3 \\
Coagulation factor X & FX & 296 & 325 & 49.6 & 5.5 \\
Farnesoid X receptor & FXR & 50 & 78 & 30.5 & 21.9 \\
Ghrelin receptor & GHSR & 73 & 66 & 100.0 & 5.8 \\
Glucocorticoid receptor & GR & 52 & 100 & 83.6 & 12.5 \\
Glycogen synthase kinase-3 $\beta$ & GSK3b & 32 & 141 & 7.5 & 6.4 \\
Histamine H1 receptor & HRH1 & 48 & 149 & 40.6 & 1.0 \\
Histamine H3 receptor & HRH3 & 239 & 335 & 36.6 & 1.6 \\
Janus kinase 1 & JAK1 & 13 & 113 & 24.6 & 0.8 \\
Janus kinase 2 & JAK2 & 34 & 163 & 20.8 & 1.0 \\
Kappa opioid receptor (agonism) & KOR (a) & 91 & 102 & 89.1 & 11.9 \\
Kappa opioid receptor (inhibition) & KOR (i) & 223 & 299 & 84.3 & 5.2 \\
$\mu$-opioid receptor & MOR & 259 & 371 & 82.1 & 6.2 \\
Orexin receptor 2 & OX2R & 160 & 137 & 79.8 & 1.3 \\
Serine/threonine-protein kinase PIM1 & PIM1 & 116 & 178 & 37.8 & 0.3 \\
Peroxisome proliferator-activated receptor $\alpha$ & PPAR$\alpha$ & 141 & 206 & 57.1 & 13.0 \\
Peroxisome proliferator-activated receptor $\delta$ & PPAR$\delta$ & 94 & 132 & 45.6 & 8.0 \\
Peroxisome proliferator-activated receptor $\gamma$ & PPAR$\gamma$ & 178 & 294 & 54.0 & 15.3 \\
Serotonin transporter & SERT & 127 & 215 & 59.6 & 5.6 \\
Sigma opioid receptor & SOR & 103 & 164 & 40.4 & 1.5 \\
PI3-kinase p110-alpha subunit & p110a & 81 & 112 & 13.0 & 0.0 \\
\bottomrule
\end{tabular}
}
\end{minipage}
\end{figure*}

\FloatBarrier

\section{Comparison models}
\label{app:comparison_models}

In our evaluation, we consider three baseline methods: MiniMol, CheMeleon, and MolFormer.
For each method, we evaluate performance using two strategies:
(1) extracting embeddings from the pre-trained backbone (without task-specific modification) and using them with a downstream predictor,
and (2) finetuning the model on each specific task.

When employing the finetuning strategy, we repeat the process over 5 random seeds for each task, using an 80/20 train/validation split.
This variation in finetuning seeds serves as a primary source of experimental variance.
Additionally, when using TabPFN as the downstream predictor (applied to embeddings from either the frozen or finetuned models),
we account for its internal variance source (\texttt{random\_state}) and the number of estimators (\texttt{n\_estimators}).
Because TabPFN splits data among estimators, the seed determines this split.
This introduces a second source of variance for finetuned methods and a primary source for frozen methods (including \minimolpp and \methodname).
These multiple sources of variance, combined with the diverse pre-training objectives of the baselines, are accounted for by the pairwise testing protocol described in the main text.

For the results reported in Table~\ref{tab:combined-results} and ~\ref{tab:ablations}, the error bars represent the \emph{mean of standard deviations}: we first calculate the standard deviation across seeds for each task, and then report the average of these per-task deviations. We chose this metric over the standard error of the mean because aggregating scores across nearly 30 tasks causes independent noise terms to average out, artificially dampening the variance. This masking effect obscures meaningful differences in stability between methods. With our chosen metric, methods with only one source of variance (e.g., \minimolpp and \methodname, which rely on fixed pretrained embeddings and only have variance from TabPFN's internal sampling) are correctly shown to have significantly lower standard deviations compared to methods involving end-to-end finetuning where variance arises from optimization, data splits, and initialization.

For each baseline we adopt the finetuning strategy and implementation provided in the CheMeleon repository. Below we provide a breakdown of the finetuning strategies.

\begin{table*}
\centering
\begin{tabularx}{\textwidth}{l >{\raggedright\arraybackslash}X >{\raggedright\arraybackslash}X >{\raggedright\arraybackslash}X}
\toprule
\textbf{Feature} & \textbf{CheMeleon} & \textbf{MiniMol} & \textbf{MolFormer} \\
\midrule
Strategy & Full finetuning & Frozen features + Head & Full finetuning \\
\midrule
Task head & chemprop default FFN & Residual MLP & Linear on CLS token \\
\midrule
Epochs & 20 & Max 50 (patience=5) & 5 \\
\midrule
Model selection & Best val loss & Best val loss & Final epoch \\
\midrule
Batch size & 64 & 64 & 8 \\
\midrule
Optimizer & Adam & Adam & AdamW \\
\midrule
Learning rate & 1e-4 & 3e-4 & 5e-5 \\
\bottomrule
\end{tabularx}
\caption{Model-specific finetuning details.}
\label{tab:finetuning-details}
\end{table*}

\end{document}